\documentclass[final,3p,times,authoryear,review,preprint]{elsarticle}

\usepackage{amssymb}
\usepackage{apalike}
\usepackage[longtable]{multirow}
\usepackage{longtable}
\usepackage{array}
\usepackage[labelfont=bf]{caption}
\usepackage{colortbl}
\usepackage{amsmath}
\usepackage{graphicx}
\usepackage{ragged2e}
\usepackage[hidelinks]{hyperref}
\hypersetup{
  colorlinks   = true, 
  urlcolor     = blue, 
  linkcolor    = blue, 
  citecolor   = blue 
}
\usepackage{multirow}
\usepackage{float}
\usepackage{subcaption} 
\usepackage{amsfonts}
\usepackage{amsmath}
\usepackage{amsthm}
\usepackage{amssymb}
\usepackage{graphicx}
\usepackage{caption} 
\usepackage{booktabs}
\usepackage{longtable} 
\usepackage{threeparttable}
\usepackage{subcaption}
\usepackage{algorithmic}
\usepackage{textcomp}
\usepackage{multicol}
\usepackage{multirow}
\usepackage{interval}
\usepackage{adjustbox}
\usepackage[hidelinks]{hyperref}
\hypersetup{
  colorlinks=true,
  urlcolor=blue,
  citecolor=blue,
  linkcolor=blue
}
\usepackage{booktabs}
\usepackage[inline]{enumitem}
\usepackage{mathrsfs}
\usepackage{float}
\usepackage{setspace}
\usepackage{tabularray}
\usepackage{xcolor}
\usepackage{rotating}
\usepackage{colortbl} 
\usepackage[linesnumbered,ruled,vlined]{algorithm2e}
\usepackage{textcomp} 
\usepackage{enumitem}
\usepackage{amsmath}
\usepackage{siunitx}
\usepackage{graphicx} 
\usepackage{tikz}
\usepackage{makecell}
\usepackage{pgfplots}
\pgfplotsset{compat=1.18}
\usepackage{pgffor}
\usepackage{ifthen}
\usepackage{diagbox}
\usepackage{listofitems}
\usetikzlibrary {arrows.meta}
\usetikzlibrary{shapes.geometric, arrows}
\usepackage{pgfplotstable}
\usepackage{ragged2e}
\usepackage{soul}
\usepackage{subcaption}

\colorlet{shadecolor}{yellow}
\usepackage{array}
\usepackage{mdwmath}
\usepackage{mdwtab}
\usepackage{eqparbox}
\usepackage{url}

\newcommand{\rh}[1]{#1}

\begin{document}

\begin{frontmatter}



\title{Optimizing Container Loading and Unloading through Dual-Cycling and Dockyard Rehandle Reduction Using a Hybrid Genetic Algorithm}


\author[inst1]{Md. Mahfuzur Rahman\corref{contrib}}
\ead{sm.mahfuz031@gmail.com}
\author[inst2]{Md Abrar Jahin\corref{contrib}} 
\ead{jahin@usc.edu}
\author[inst1]{Md. Saiful Islam}
\ead{saifuliem@iem.kuet.ac.bd}
\author[inst3]{M. F. Mridha}
\ead{firoz.mridha@aiub.edu}

\affiliation[inst1]{organization={Department of Industrial Engineering and Management},
    addressline={Khulna University of Engineering \& Technology (KUET)}, 
    city={Khulna},
   postcode={9203}, 
   country={Bangladesh}}
\affiliation[inst2]{organization={Thomas Lord Department of Computer Science},
    addressline={Viterbi School of Engineering, University of Southern California (USC)},
    city={Los Angeles},
    state={CA},
    postcode={90089},
    country={USA}}
\affiliation[inst3]{organization={Department of Computer Science},
    addressline={American International University-Bangladesh (AIUB)}, 
    city={Dhaka},
    postcode={1229},
    country={Bangladesh}}

\cortext[corauthor]{Corresponding author}
\cortext[contrib]{Authors contributed equally}

\begin{abstract}
This paper addresses the NP-hard problem of optimizing container handling at ports by integrating Quay Crane Dual-Cycling (QCDC) and dockyard rehandle minimization. We identify critical inter-dependencies between the unloading sequence of QCDC and the dockyard container arrangement and propose the \ul{Q}uay \ul{C}rane \ul{D}ual \ul{C}ycle -- \ul{D}ockyard \ul{R}ehandle \ul{G}enetic \ul{A}lgorithm (QCDC-DR-GA), a hybrid Genetic Algorithm (GA) that holistically optimizes both aspects jointly: maximizes the number of Dual Cycles (DCs) and minimizes the number of dockyard rehandles. QCDC-DR-GA employs a mixed 1D--2D chromosome representation with specialized crossover and mutation operators tailored to each component. Extensive experiments across six scenarios spanning small, medium, and large container ships demonstrate that QCDC-DR-GA reduces total operation time by up to 30.1\% compared to existing methods (20.1\% on average across large-ship scenarios). Statistical validation via two-tailed paired t-tests confirms significant improvements in 20 out of 24 pairwise comparisons at the 5\% significance level ($\alpha = 0.05$, $t_{19}(0.05) = 2.093$). The results underscore the inefficiency of isolated optimization and highlight the critical need for integrated algorithms in port operations. This approach increases resource utilization and operational efficiency, offering a cost-effective solution for ports to decrease turnaround times without infrastructure investments.
\end{abstract}



\end{frontmatter}



\section{Introduction}
Maritime trade is a cornerstone of the global economy, with seaports serving as critical nodes in international logistics networks. {Approximately 80\% of global merchandise trade is transported by sea ~\citep{80_pctg_world_trade_by_sea}. Although a substantial share of seaborne cargo tonnage consists of bulk and tanker shipments, containerized shipping plays a pivotal role in global supply chains by carrying most manufactured and other higher-value goods through scheduled liner services.} Consequently, maritime vessels play a vital role in sustaining this system, prompting many nations to invest in expanding and modernizing their shipping fleets. Contemporary ultra-large container vessels (ULCVs) are capable of carrying more than 25,000 TEUs (twenty-foot equivalent units)~\citep{ship_size}. Accommodating these vessels requires ports to undertake significant operational and logistical preparations. One of the primary performance indicators for port efficiency is the turnaround time of vessels. Fifteen years ago, the average turnaround time for a ULCV was approximately 14–15 days; today, it has been reduced to 3–4 days. Future targets aim to further decrease this time to a matter of hours. Over the years, shipping companies have implemented numerous operational strategies and technical optimizations to accelerate the loading and unloading processes. This study proposes a novel approach that contributes to this ongoing effort.

Among container handling equipment, Quay Cranes (QCs) are the most capital-intensive asset in terminal operations. So, the availability and utilization of QCs frequently constitute a critical bottleneck in port performance \citep{chu2002aggregates}. Improving QC efficiency has the potential to significantly reduce vessel turnaround times, increase terminal productivity, and improve the overall throughput of the freight transportation system \citep{goodchild2006double}. This research targets this key operational constraint and presents a low-cost, implementable strategy to increase QC productivity. Importantly, the proposed solution does not require the development of new infrastructure or the integration of advanced technologies. While it may not resolve long-term capacity limitations, it provides a rapid deployment timeline compared to infrastructure-intensive approaches, such as optimizing berth and crane allocations \citep{tuncel2024optimization}, determining the required number of automated guided vehicles (AGVs) \citep{li2024integrated}, or balancing trade-offs between storage capacity and handling workload \citep{luo2011storage}. Moreover, the proposed method can be effectively integrated with such long-term solutions to yield synergistic benefits.

\begin{figure}[htbp]
\centering
\includegraphics[width=1\linewidth]{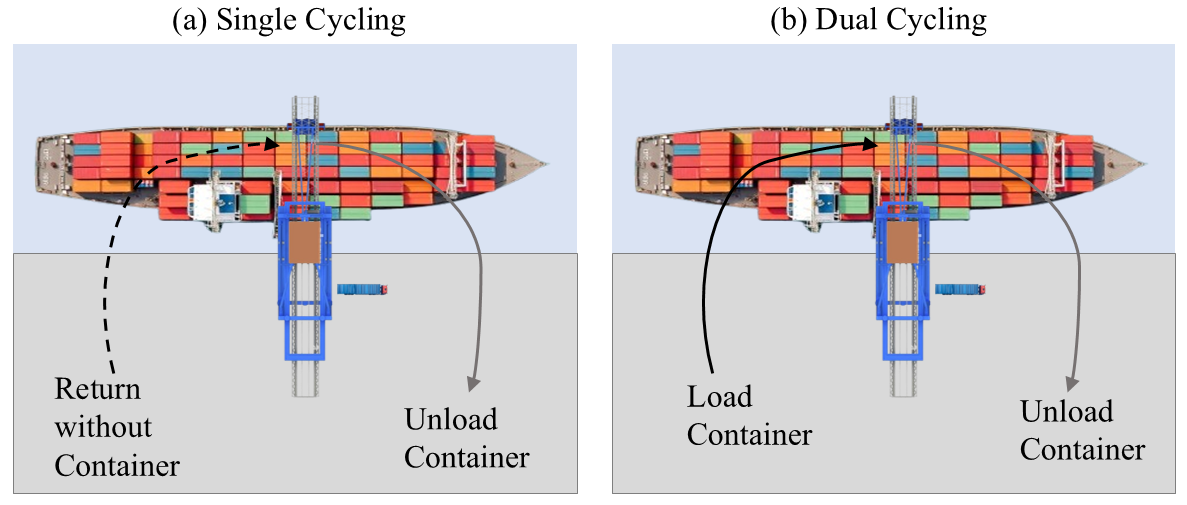}
\caption{(a) Unloading using single cycling; (b) Simultaneous unloading and loading using double cycling.}
\label{fig:single_and_dual_cycling}
\end{figure}

Two significant challenges in QC operations are central to this work. The first involves the operational strategy for the QC itself. Most container terminals currently use the \textit{Single Cycle} (SC) strategy for QC operations, in which unloading is completed before the loading phase begins. In this method, each operational cycle includes an empty return movement of the QC in each cycle (Figure~\ref{fig:single_and_dual_cycling}a). The \textit{Dual Cycling} (DC) approach presents a more efficient alternative for QC operations by allowing simultaneous loading and unloading. In this method, a container designated for loading is positioned and ready while the crane is engaged in unloading. Upon completion of the unloading task, the QC immediately proceeds to load the prepared container, thereby eliminating the need for an idle return movement (see Figure~\ref{fig:single_and_dual_cycling}b). This simultaneous operation effectively increases QC productivity and contributes to reducing vessel turnaround time. As shown by \citet{goodchild2006double}, the sequence in which containers are unloaded from bay stacks\footnote{Definitions related to bay stacks are provided in Section~\ref{problem}. See also Figure~\ref{fig:row_tier_bay}.} significantly affects the number of achievable dual cycles. They also proposed a greedy heuristic algorithm to identify near-optimal unloading sequences in real time, aiming to maximize DC opportunities. Other studies confirm that DC can reduce QC operational time by 10–20\% and is a key efficiency strategy in contemporary terminals \citep{chu2017multiple}. Building on this foundational work, subsequent research has focused on improving the dual cycling strategy using a variety of optimization techniques. The Quay Crane Dual-Cycling Scheduling (QCDCS) problem, as modeled in the same study, can be formulated as a two-machine flow shop scheduling problem. Due to its combinatorial complexity, it is considered NP-hard (see Section~\ref{complexity}). Consequently, metaheuristic approaches such as Genetic Algorithms (GA) have been widely implemented to obtain high-quality solutions within acceptable computational timeframes (see Section~\ref{solution_approach}).

The second major challenge is rehandling during loading and unloading. Rehandling is necessary when retrieving a specific container, the target container, that is not at the top of its stack. To access the target container, any obstructing containers must be relocated, which may, in turn, lead to additional rehandling during subsequent retrieval operations \citep{sauri2011space}. The decision on where to temporarily relocate a blocked container, typically based on criteria such as proximity, stack height, and the selected optimization strategy, can significantly affect operational efficiency. Frequently, the obstructed container is relocated to another stack within the same yard bay, potentially introducing further complications. Given that rehandling operations are both time- and resource-intensive, port terminals try to minimize their frequency. The problem of minimizing rehandles has also been shown to be NP-hard, and, as before, metaheuristic techniques, particularly GA, have proven effective in producing high-quality solutions within practical time constraints.

Recent literature has increasingly focused on integrated optimization for port operations. \rh{Throughout this paper, we use the term \emph{integrated optimization} to denote approaches that bring multiple terminal sub-processes (e.g., berth allocation, crane scheduling, yard planning) into a single coordinated planning framework, as opposed to \emph{independent} (or \emph{isolated}) optimization, in which each sub-process is solved separately, one at a time, with the output of one stage passed as a fixed input to the next. We further distinguish a stronger notion that is the specific target of this work, namely \emph{joint optimization}: the simultaneous optimization of two mutually dependent decisions, the dual-cycling unloading sequence ($A_1$) and the dockyard configuration ($A_2$), within a single objective and a single search loop, so that trade-offs between them are resolved endogenously rather than by sequential hand-off. Integrated optimization is therefore a broad umbrella; joint optimization is the specific, tighter form of integration we pursue.} For example, \citet{xia_loading_2020} implemented a Double Deep Q-Network (DDQN) for container loading, while \citet{yan_optimizing_2024} introduced a Deep Reinforcement Learning (DRL) framework for container repositioning. Other works have proposed tailored heuristics, such as branch-and-bound algorithms for minimizing crane operation time~\citep{parreno-torres_minimizing_2020}, a Probabilistic Genetic Algorithm (PGA) for integrated terminal scheduling~\citep{lu_three-stage_2021}, and a two-stage GA for job instruction sequencing~\citep{fibrianto_job_2020}. \rh{However, even these works that are described as ``integrated'' typically optimize their constituent sub-problems in a sequential or loosely coupled manner: a decision is made for one component (e.g., the crane schedule) and then treated as fixed while a second component (e.g., the yard plan) is optimized. In other words, the broad goal of integration has been pursued, but the specific coupling between the dual-cycling sequence and the dockyard layout has remained an independent, stage-wise process.} Indeed, these advanced methods typically address individual components of port optimization, such as crane scheduling or rehandle reduction, in isolation.

In contrast, this research addresses the critical interdependence of these two objectives \rh{through true joint optimization rather than stage-wise integration}. We observed that an unloading sequence designed to maximize dual cycles can inadvertently increase the number of dockyard rehandles, and vice versa. This conflict arises because dual cycling requires a specific ordering of stack unloading operations that minimizes crane idle movements; however, this ordering simultaneously determines the sequence in which loading containers are demanded from the dockyard. If the dockyard arrangement was optimized independently of the unloading sequence, as in all prior work, then the demanded loading sequence may not match the dockyard layout, leading to additional rehandles.

To illustrate why conventional independent optimization is inadequate, consider the following: a port planner who uses a DC-focused algorithm to determine the unloading sequence $A_1$ will then hand this sequence to a separate dockyard management system to arrange containers in $A_2$. However, the dockyard system has no knowledge of the unloading sequence constraints imposed by dual cycling, and the DC algorithm has no knowledge of the rehandle implications of its chosen sequence. In practice, optimizing these two components in isolation creates a feedback gap: improving one metric often degrades the other. The only way to resolve this conflict is to optimize both jointly within a single algorithmic loop, allowing the optimizer to make explicit trade-offs between maximizing dual cycles and minimizing rehandles. This is precisely the motivation for our integrated design.

Furthermore, the rapid advancement of computational resources in recent years makes this joint optimization feasible in near-real-time port planning contexts. While organizational barriers between quay crane scheduling departments and yard management departments have historically hindered integration, the growing adoption of Terminal Operating Systems (TOS) that unify vessel and yard data provides an ideal deployment environment for algorithms like QCDC-DR-GA. Our method is designed to be invoked as a pre-arrival planning module within such systems, producing both the optimal unloading sequence and the optimal dockyard configuration simultaneously before the vessel berths.

This work aims to resolve this conflict by optimizing the entire process within a single, integrated algorithm. We propose a method called \textit{Maximizing QCDCs and Minimizing Dockyard Rehandles by GA} (QCDC-DR-GA) to produce an unloading sequence and a dockyard container arrangement that simultaneously maximizes dual cycles and minimizes rehandles. The algorithm's robustness is verified by comparing it with results from strategies that address these goals separately: the QCDCS Greedy Upper Bound, GA-QCDCS, GA-ILSRS with single cycling, and GA-ILSRS with dual cycling.

This study presents seven significant contributions:
\begin{enumerate}
\item This study empirically validates the correlation between the unloading sequence of stacks and the occurrence of dockyard rehandles, particularly in the context of dual-cycling strategies.
\item We develop a comprehensive algorithm that integrates dockyard and QCDC operations to holistically optimize container handling.
\item This paper introduces a novel hybrid GA to efficiently solve the proposed integrated problem.
\item We propose a specialized GA that combines one- and two-dimensional techniques tailored to container-handling challenges.
\item We conduct an extensive analysis to identify the most influential computational parameters and methods within the GA framework.
\item We provide a clear rationale for the strategies and methodologies selected at various stages of the optimization process.
\item We perform a comprehensive benchmarking analysis, validating the superiority of our proposed strategy against four state-of-the-art algorithms through statistical paired t-tests.
\end{enumerate}

This article is structured as follows: The ``\hyperref[literature]{Literature Review}'' section discusses relevant work on DC and rehandling. The ``\hyperref[problem]{Problem Description}'' section outlines the problem statement. The ``\hyperref[method]{Methodology}'' section details the algorithm formulation, objectives, constraints, and our proposed QCDC-DR-GA approach, including its workflow, strategies, and parameters. The ``\hyperref[results]{Results}'' section covers scenario generation, computational experiments, and result analysis. The ``\hyperref[discussion]{Discussion}'' discusses the practical and managerial implications and the unique theoretical contributions of the proposed approach. Finally, the ``\hyperref[conclusion]{Conclusions}'' section summarizes the work, highlights its contributions, and suggests future research directions.

\section{Literature Review}
\label{literature}

A considerable volume of operational research has focused on addressing challenges in the port sector. These studies commonly concentrate on matters related to strategic design and planning, such as the optimal number of berths and crane combinations \citep{aslam2024survey}, the optimal size of storage space \citep{tuncel2024optimization}, determination of the number of AGVs required \citep{li2024integrated}, and trade-offs between storage space and handling work \citep{luo2011storage}. Also, work has been done on operational scheduling in berths and QCs \citep{ech2022berth}, dispatching methods for yard cranes and AGVs \citep{zhou2023joint}, and dynamic deployment and scheduling of yard cranes \citep{chen2007tabu, GUO2011472}.

Since QCs are the main bottleneck in the efficient operation of container terminals, their operational efficiency determines the turnaround time of ships in seaports \citep{chu2002aggregates}. Hence, more work has been done to improve the efficiency of the QC operation. \cite{kim2004crane} formulated a mixed-integer programming algorithm that accounted for diverse constraints concerning QC operations. They introduced a heuristic search algorithm named "greedy randomized adaptive search procedure" to address this issue effectively. \cite{tavakkoli2009efficient} introduced a novel mixed-integer programming (MIP) algorithm for solving the Quay Crane Scheduling and Assignment problem (QCSAP), as traditional methods struggle to solve it efficiently within reasonable timeframes. To address this, they presented a GA as a solution for real-world scenarios. Also, \cite{fu2014multi}, along with \cite{diabat2014integrated}, addressed the combination of QC assignment and scheduling problems. They also used the GA to find solutions. Their results showed that the GA was faster and more effective than the Lagrangian relaxation technique.

The effectiveness of QCs depends on how well they work with other equipment, such as yard trailers and cranes. Certain researchers investigated solving the integration scheduling problem to improve coordination and overall efficiency at container terminals. For example, \cite{bish2003multiple} introduced algorithms that combined various sub-processes. These included finding suitable storage locations for unloaded containers, dispatching vehicles to specific containers, and planning loading and unloading activities for quay and yard cranes. \cite{chen2007tabu} created a comprehensive algorithm to increase the efficiency of the complete loading and unloading procedure. \cite{zeng2011disruption} created an algorithm for recovering from berth and QC schedule disruptions. However, the previously mentioned literature on scheduling primarily focuses on the Quay Crane Single Cycling (QCSC) method, with relatively limited advancement in QCDCS scheduling.

\subsection{Dual Cycling}\label{dualcycle}
\cite{goodchild2006double} first introduced the QCDCS strategy. They formulated the double-cycling problem as a 2-machine flow shop problem and proposed a greedy approach to generate a loading and unloading sequence. They also ran a trial at the Port of Tacoma, US (2003). In this trial, the revised mean time for an SC was 1 minute and 45 seconds, and for a double cycle, 2 minutes and 50 seconds. Consequently, double cycling resulted in a time savings of 40 seconds per pair of containers subjected to the process. Later, they devised a framework for assessing QC performance \citep{goodchild2007crane}. They also formulated a straightforward formula to forecast the effect on turnaround time. Their findings also revealed that adopting a double-cycling approach could reduce operating time by 10\% and decrease demand for yard tractors and drivers.  \cite{song2007study} proposed a formula to find the optimal starting point of double cycling that maximizes its frequency.

The above studies concentrated on implementing QCDCS for single QCs. Their practical experiments revealed that this approach could increase each QC's productivity by 10-20\%. Then \cite{zhang2009maximizing} proposed multiple QCDC algorithms and solved a mixed-integer programming problem. At the same time, the sequence was generated using a constructive Johnson’s rule with an effective local search method. \cite{ku2016double} pointed out a flaw with the existing multiple QCDCS algorithm that lets cycles that are not implementable.

\cite{lee2015optimal} investigated the computational complexity of the QCDCS problem. They showed that it can be formulated as a flow-shop scheduling problem with series-parallel precedence constraints and solved it in polynomial time. For ease of implementation, they presented an optimal algorithm for the general QCDCS problem, a simplified version of Sidney’s algorithm. \cite{zeng2015simulation} developed a mixed-integer programming algorithm for QCDCS. The algorithm considered the stowage plan of outbound containers and the operational sequence of QC. A heuristic method called bi-level GA was designed to solve the algorithm. \cite{zhang2019modeling} focused on overall handling efficiency and the system’s stability of container terminals with double cycling. \cite{he2020quay} solved a mixed integer programming model, which covers the main operational constraints (including multiple hatch-covers) in a container terminal. \cite{zheng2020dynamic} presented a dynamic programming approach to solve the QCDCS problem optimally. \cite{ahmed2021synchronized} developed two simulation algorithms and implemented them based on a real-life case study, considering uncertainties in the work task duration. 

\subsection{Reducing Dockyard Rehandles}
The issue of rehandling is typically connected to the arrangement of containers on ships and in yards. Some studies have addressed the rehandling problem by examining container ship stowage plans or yard storage arrangements. They focused on refining the rehandling strategy based on specific loading sequences, without considering how those sequences might affect rehandling. \cite{kim1997evaluation} proposed a methodology to calculate the expected number of rehandles to pick up a random container and the total number of rehandles to pick up all the containers in a bay for a given initial stacking configuration. \cite{imai2006multi} formulated the issue as a multi-objective integer programming challenge. They employed the weighting method and acquired a collection of non-inferior solutions. \cite{sauri2011space} suggested three stacking strategies, which consider the containers’ arrival rates, departure rates, and storage yard characteristics, and then developed an algorithm to determine the number of rehandles. \cite{lee2009heuristic} proposed a novel approach that integrated yard truck scheduling and storage allocation and developed a hybrid insertion algorithm to solve this problem. \cite{caserta2011applying} and \citep{wang2024optimization} presented a yard crane scheduling problem to carry out a set of container storage and introduced an algorithm based on the corridor method designed to address the problem of relocating blocks within block stacking systems.

Various algorithms are used to address relocation problems, but the GA is commonly adopted due to its efficiency. For example, \cite{homayouni2014genetic} integrated the scheduling of QCs, AGVs, and handling platforms and proposed a GA to solve the problem. \cite{ji2015optimization} presented an improved GA design to solve the algorithm. Compared with earlier strategies and heuristics, they demonstrated the effectiveness of their optimization approach and algorithm. 

\subsection{Summary}
Table \ref{tab:literature_sum} summarizes the reviewed literature. This literature highlights optimizing port operations by focusing on strategic planning, QC efficiency, integrated scheduling with other terminal equipment, and advanced techniques such as dual cycling. Various algorithms, including mixed-integer programming and heuristic algorithms, have been proposed to improve crane coordination, reduce turnaround time, and minimize container rehandling at the dockyard. GAs have recently been used due to their effectiveness in solving complex scheduling and relocation problems.

\begin{table*}[!ht]
\centering
\caption{Literature summary and analysis.}
\label{tab:literature_sum}
\begin{tabular}{p{0.25\columnwidth}p{.35\columnwidth}p{.35\columnwidth}}
\toprule[1.5pt]
\textbf{Type} & \textbf{Literature} & \textbf{Key Feature}  \\ 
\midrule[1pt]
Different port operations & [\cite{aslam2024survey}, \cite{tuncel2024optimization}, \cite{li2024integrated}, \cite{luo2011storage}, \citep{ech2022berth}, \cite{zhou2023joint}, \cite{chen2007tabu}, \citep{GUO2011472}] &  Focuses on optimizing port operations, including strategic planning of resources like berths, cranes, and storage space, as well as scheduling and deployment of handling equipment such as AGVs and yard cranes.\\

QC Scheduling and Optimization & [\cite{chu2002aggregates}, \cite{kim2004crane}, \cite{tavakkoli2009efficient}, \cite{fu2014multi}, \cite{diabat2014integrated}] & Aimed to increase QC efficiency. Their approaches significantly reduced computation time and improved solution quality compared to traditional methods.\\

Integrated Scheduling & [\cite{bish2003multiple}, \cite{chen2007tabu}, \cite{zeng2011disruption}] &  Addressed tasks like container storage allocation, equipment dispatching, and disruption recovery to improve overall operational efficiency.\\

QCDC (different greedy approaches, complexity analysis and case Studies) & [\cite{goodchild2006double}, \cite{goodchild2007crane}, \cite{song2007study}, \cite{zhang2009maximizing}, \cite{ku2016double}, \cite{lee2015optimal}, \cite{ahmed2021synchronized}] &  Developed optimization algorithms, practical frameworks, and greedy algorithms that demonstrated up to 20\% efficiency gains, while also identifying limitations in existing multi-crane algorithms.\\

QCDC (Integer Programming) & [\cite{he2020quay}, \cite{zeng2015simulation}] &  Covers the main operational constraints (including multiple hatch-covers).\\

QCDC (Dynamic programming) & [\cite{zheng2020dynamic}] & Solved the problem optimally.\\

QCDC (GA) & [\cite{zeng2015simulation}] & Bi-level GA was designed to solve the algorithm\\

Dockyard rehandles (Different greedy approaches) & [\cite{kim1997evaluation}, \cite{imai2006multi}, \cite{sauri2011space}, \cite{lee2009heuristic}, \cite{caserta2011applying}, \cite{wang2024optimization}] & Approaches include mathematical modeling, scheduling integration, and heuristic algorithms, focusing on factors like loading sequences, container arrival and departure rates, and storage configurations. \\

Dockyard rehandles (GA) & [\cite{homayouni2014genetic}, \cite{ji2015optimization}] & Proposed improved GAs to optimize dockyard rehandles.\\
\bottomrule[1.5pt]
\end{tabular}
\end{table*}

Previous research separately addressed maximizing DCs in the container loading-unloading system through QC and minimizing dockyard rehandles. This work identifies that two scenarios (loading-unloading by QCDC and dockyard rehandles) are interdependent. There is an optimal unloading-loading sequence within the dockyard plan for which the total unloading and loading time is minimum. Our QCDC-DR-GA algorithm finds the ideal unloading sequence and dockyard layout to maximize DCs and minimize rehandles. Proven effective against existing methods, it offers a holistic solution for efficient container handling at ports.

\section{Problem Description}
\label{problem}
The layout of containers on a ship or within a port yard, as illustrated in Figure~\ref{fig:row_tier_bay}, can be represented as a three-dimensional matrix. {The four fundamental structural units used to describe container arrangements on ships and in yards are defined as follows:

\begin{enumerate}
    \item \textbf{Tier}: The vertical layer at which a container sits within a stack. Tier~1 is the ground level (bottom of the stack); containers are stacked upward in increasing tier numbers.
    \item \textbf{Row}: A transverse line of stacks running across the full width of the ship (or yard bay). Each row is processed as an independent unit by a quay crane.
    \item \textbf{Bay}: A longitudinal section of the ship running along its length. A bay contains one or more transverse rows of stacks.
    \item \textbf{Stack}: A vertical column of containers occupying a single ground-level position within a row and bay. A stack is the fundamental handling unit in this model; the quay crane unloads containers from the top of a stack downward and loads them upward. In the dockyard, a stack similarly refers to a vertical column of containers at a fixed yard ground position. The height of a stack is bounded by the maximum permitted tier count $H_{mn}$ (Constraint~\eqref{eq:con7}).
\end{enumerate}
}

Containers are stacked vertically into tiers, arranged side-by-side into stacks within a row, and aligned along the ship’s length into bays. A row typically spans the full width of the bay or the vessel. Modern container vessels are capable of accommodating up to 30 container stacks per row and up to 30 rows along the ship’s length, particularly when transporting 40-foot containers. Container ships vary widely in size and type. In this study, we adopt the classification proposed by \citet{ship_size}, categorizing ships into three distinct types based on their container-carrying capacity, as presented in Table~\ref{tab:ship_sizes}. The carrying capacity of these vessels is expressed in Twenty-foot Equivalent Units (TEUs), where a standard 20-foot container equals 1 TEU, and a 40-foot container is equivalent to 2 TEUs.

\begin{table}[!ht]
\caption{Classification of ship types based on container capacity. The table lists the number of containers that small, medium, and large ships can carry. This classification is used to define ship categories in the proposed algorithm.}
\label{tab:ship_sizes}
\centering
\begin{tabular}{c c c}
    \toprule[1.5pt]
    \textbf{Ship Size} & \textbf{No. of Containers (2 TEU)} & \textbf{No. of Containers (1 TEU)} \\
    \midrule[1pt]
    \multirow{2}{*}{Small} & \multicolumn{1}{c}{2,500} & \multicolumn{1}{c}{5,000}\\
     & \multicolumn{1}{c}{3,500} & \multicolumn{1}{c}{7,000}\\
    \midrule[0.8pt]
     \multirow{2}{*}{Medium} & \multicolumn{1}{c}{6,500} & \multicolumn{1}{c}{13,000}\\
     & \multicolumn{1}{c}{8,000} & \multicolumn{1}{c}{16,000}\\
     \midrule[0.8pt]
    \multirow{2}{*}{Large} & \multicolumn{1}{c}{10,000} & \multicolumn{1}{c}{20,000}\\
    & \multicolumn{1}{c}{12,000} & \multicolumn{1}{c}{24,000}\\
    \bottomrule[1.5pt]
\end{tabular}

\end{table}

QC usually processes unloading and loading across the ship bay. In the QCSC method, loading operations can only be performed after unloading operations are complete. However, in the QCDC method, a QC performs unloading and loading operations simultaneously in a particular ship bay. 

\begin{figure}[!ht]
\tikzset{
    box1/.style={rectangle, minimum size=5mm, text centered, draw=black!90, fill=green!70, thick},
    box2/.style={rectangle, minimum width=15mm, minimum height=5mm, draw=black!90, fill=green!50, thick},
    box3/.style={rectangle, minimum width=3mm, minimum height=3mm, draw=black!70, fill=black!40}
}
\tikzstyle{arrow} = [line width=0.3mm, -{Stealth[length=3mm, round]}]
\tikzstyle{two_sided} = [line width=0.3mm, <->]

\begin{center}
\begin{tikzpicture}
    \foreach \i in {-3,...,3} {
        \node (\i) [box2] at (-36mm, \i mm * 5) {};
    }
    \foreach \i in {-3,...,3} {
        \node (\i) [box2] at (-18mm, \i mm * 5) {};
    }
    \foreach \i in {-3,...,3} {
        \node (\i) [box2] at (0mm, \i mm * 5) {};
    }
    \foreach \i in {-3,...,3} {
        \node (\i) [box2] at (18mm, \i mm * 5) {};
    }
    \draw[black, very thick] (-44mm, -2.5mm) -- (29mm, -2.5mm);
    \draw[black, very thick] (-44mm, -2.5mm) -- (-46mm, 2mm);
    \draw[black, very thick] (-61mm, 2mm) -- (-46mm, 2mm);
    \draw[black, very thick] (-61mm, 2mm) -- (-51mm, -25mm);
    \draw[black, very thick] (-51mm, -25mm) -- (-58mm, -25mm);
    \draw[black, very thick] (-58mm, -25mm) arc[start angle=90, end angle=270, radius=3mm];
    \draw[black, very thick] (-58mm, -31mm) -- (29mm, -31mm);
    \node (top_view) at (-1, -35mm) {(a) Side view};
    \node (front_view) at (7.5, -35mm) {(b) Front view};
    \foreach \i in {10,...,20} {
        \foreach \j in {-3,...,3}{
            \node (\i) [box1] at (\i mm * 5, \j mm * 5) {};
        }
    }
    \draw[black, very thick] (46mm, 3mm)--(104mm, 3mm);
    \draw[black, very thick] (46mm, 3mm) -- (46mm, -25mm);
    \draw[black, very thick] (104mm, 3mm) -- (104mm, -25mm);
    \draw[black, very thick] (46mm, -25mm) arc[start angle=180, end angle=270, radius=5mm];
    \draw[black, very thick] (104mm, -25mm) arc[start angle=0, end angle=-90, radius=5mm];
    \draw[black, very thick] (51mm, -30mm)--(99mm, -30mm);
    \draw[two_sided] (45mm, 20mm) -- (105mm, 20mm) node[midway, above]{Rows};
    \draw[black] (55mm, 17.5mm)--(55mm, 25mm);
    \draw[black] (95mm, 17.5mm)--(95mm, 25mm);
    \draw[black] (50mm, 25mm) rectangle (100mm, 31mm);
    \foreach \i in {1,...,8} {
        \node (\i) [box3] at (\i * 5mm + 52mm, 28mm) {};
    }
    \draw[two_sided] (-46mm, 20mm) -- (28mm, 20mm) node[midway, above]{Bays};
    \draw[two_sided] (29mm, 20mm) -- (29mm, -20mm) node[midway, right]{Tiers};
\end{tikzpicture}
\end{center}
    \caption{Container arrangement on a ship, illustrating the three-dimensional storage structure. \rh{(a)~Side view, taken along the ship's length, showing \emph{bays} and \emph{tiers}; (b)~front view, taken across the ship's width, showing \emph{rows} and \emph{tiers}. A \emph{tier} is the vertical level (Tier~1 at the bottom, increasing upward), a \emph{stack} is a single vertical column of containers at one ground position, a \emph{row} is the transverse line of stacks across the width, and a \emph{bay} is a longitudinal section along the length.}}
    \label{fig:row_tier_bay}
\end{figure}

The operating cycle of a QC can be broken down into the following parts: 
\begin{enumerate}[label=(\roman*)]
    \item Locking and unlocking the trolley with the container. 
    \item Horizontal movement of the trolley (with container). 
    \item Vertical movement of the trolley (with container). 
\end{enumerate}

\subsection{Case Consideration}
After the arrival of a ship in port with a set of containers to be unloaded and a loading plan for a set of containers, let $U_{c}$ and $L_{c}$ be the numbers of containers to be unloaded and loaded, where $c$ is the stack number. Figure \ref{fig:cont_plan} illustrates an example used in this work. 

Let $S$ be the set of stacks in a row. $|s|=N$ denotes the number of stacks in the set $S$, and $P$ a permutation of set $S$ telling the ordering of the stacks. \rh{Throughout this paper, we adopt a single, consistent notation for the three spatial identifiers, which we fix here once and use everywhere (in the text, figures, tables, and in $A_1$ and $A_2$):
\begin{itemize}
    \item \textbf{Stack identifier:} stacks in a row are indexed by the integers $1,2,\dots,N$ from one side of the row to the other. When a purely illustrative label is needed in a small schematic, we may equivalently letter them $A,B,C,\dots$; in that case, the letter is only a display alias for the stack index ($A\!\equiv\!1$, $B\!\equiv\!2$, \dots). The numeric form is the canonical one and is the form used inside $A_1$.
    \item \textbf{Tier identifier:} tiers are numbered $1,2,\dots$ from the bottom of a stack upward, so Tier~1 is always the ground level. For compact cell labels we use the alphabetic tier aliases $\mathrm{A}\!\equiv\!\text{Tier }1$, $\mathrm{B}\!\equiv\!\text{Tier }2$, $\mathrm{C}\!\equiv\!\text{Tier }3$, and so on; the alphabetic and numeric tier forms are interchangeable and both count upward from the bottom.
    \item \textbf{Container identifier:} a container is named by concatenating its destination \emph{ship-stack number} and its \emph{tier}, written \texttt{[stack][tier]}. For example, `\texttt{2C}' denotes the container destined for ship stack~2, tier~3 (i.e., tier alias~C). This single \texttt{[stack][tier]} convention is used in Figure~\ref{fig:cont_plan}, Tables~\ref{table3}--\ref{table2}, and in the dockyard plan $A_2$.
\end{itemize}}
For example, in Figure \ref{fig:cont_plan} (a), the set of stacks is $S=\{A, B, C, D\}$ \rh{(equivalently $\{1,2,3,4\}$ under the alias above)}. Then $P(1) = A$, $P(2) = B$, $P(3) = C$, and $P(4) = D$. The order in which the stacks within each row are handled affects the total number of cycles \citep{FAZI2023343,goodchild2006double}, and there is an optimum sequence of the set $S$ for which the number of cycles $w$ is minimum. Other research supports the significant impact of stack sequence on performance, noting that optimizing reordering and retrieval minimizes unproductive moves \citep{ji2015optimization}, and sequence decisions directly affect the total number of container relocations and cycles \citep{DAYAMA2017307,kuznetsov2020optimization}.

\subsubsection{Generic Double Cycling Method}
\begin{enumerate}[label=(\roman*)]
    \item Select any unloading permutation, $P'$. Unload all the containers from the first stack, then the second, and continue until all stacks have been unloaded.  
    \item Select a loading permutation, $P$, and load the stacks according to that permutation. Loading can be started in any stack when it becomes empty or contains only containers that would not be unloaded at this port. Once loading has been initiated on a stack, continue the process until that stack is fully loaded. 
\end{enumerate}

\begin{figure}[!ht]
    \centering
\newcommand{\co}{{b, b, 1C, 1B, 1A}, {2C, 2B, 2A}, {b, b, 3B, 3A}, {b, b, 4B, 4A}}

\newcommand{\coo}{{b, b, 1A, 1B, 1C}, {2A, 2B, 2C, 2D}, {b, b, 3A}, {b, b, 4A, 4B, 4C}}

\newcommand{\cooo}{{b, b, 2A, 2B}, {b, 2C, 4A}, {b, 1A, 1B, 3A}, {b, b, 4C, 4B, 2D}}
\tikzset{
    box1/.style={rectangle, minimum size=8mm, text centered, draw=black!40, fill=black!10, very thick},
    box2/.style={rectangle, minimum size = 8mm, text centered, draw=black!100, fill=black!80, very thick},
    box3/.style={rectangle, minimum size=8mm, text centered, draw=black!50, fill=black!30, very thick},
    box4/.style={rectangle, minimum size=8mm, text centered, draw=blue!50, fill=blue!10, very thick},
    arrow/.style={very thick, ->}
}

\begin{center}  
\begin{tikzpicture}
    \newcounter{r}
    \newcounter{c}
    \setcounter{c}{0}

    \foreach \col in \co {
        \setcounter{r}{0}
        \foreach \i in \col {
            \ifthenelse{\equal{\i}{b}} {
                \node (\i) [box2] at (\value{c}, \value{r}) {};
            } {
                \node (\i) [box1] at (\value{c}, \value{r}) {\i};
            }
            \stepcounter{r}
        }
        \stepcounter{c}
    }
    \newcommand{\name}{A, B, C, D}
    \setcounter{c}{0}
    \foreach \i in \name {
        \node (\i) at (\value{c}, -0.7) {\i};
        \stepcounter{c}
    }
    \node (a1) [box3] at (0, 4) {1A};
    \node (a1) [box3] at (3, 3) {4A};

    \draw[black, thick] (4.4, 0.4) rectangle (11.2, 4.3);
    \node[font=\small\bfseries] at (7.8, 3.9) {Legend};    
    \node (1) [box1] at (5, 3){};
    \node (2) [box3] at (5, 2){};
    \node (3) [box2] at (5, 1){};
    \node (1k) at (7, 3) {Container to unload};
    \node (2k) at (7.5, 2) {Container to be rehandled};
    \node (3k) at (7.6, 1) {Container to stay on vessel};
    \node (cap) at (4.2, -1.3) {(a) Unloading-plan of a vessel};

    \setcounter{c}{0}
    \foreach \col in \coo {
        \setcounter{r}{-7}
        \foreach \i in \col {
            \ifthenelse{\equal{\i}{b}} {
                \node (\i) [box2] at (\value{c}, \value{r}) {};
            } {
                \node (\i) [box4] at (\value{c}, \value{r}) {\i};
            }
            \stepcounter{r}
        }
        \stepcounter{c}
    }
    \setcounter{c}{0}
    \foreach \i in \name {
        \node (\i) at (\value{c}, -7.7) {\i};
        \stepcounter{c}
    }
    \node (cap1) at (1.5, -8.3) {Ship Loading Sequence};
    \node (cap2) at (6.6, -8.3) {Dockyard container's status};
    \node (cap3) at (4.2, -9) {(b) Loading plan of the vessel};
    
    \setcounter{c}{5}
    \foreach \col in \cooo {
        \setcounter{r}{-7}
        \foreach \i in \col {
            \ifthenelse{\equal{\i}{b}} {
                \node (\i) [box2] at (\value{c}, \value{r}) {};
            } {
                \node (\i) [box4] at (\value{c}, \value{r}) {\i};
            }
            \stepcounter{r}
        }
        \stepcounter{c}
    }
\end{tikzpicture}
\end{center}
    \caption{\rh{Illustration of the unloading and loading plan of a single ship row, drawn with tiers increasing upward (Tier~1 at the bottom of each column) and stacks placed side by side along the horizontal axis. Containers are labeled with the \texttt{[stack][tier]} convention of Section~\ref{problem}; the stack labels $A,B,C,D$ beneath the columns are display aliases of stack indices $1,2,3,4$. (a)~The unloading plan; (b)~the loading plan, whose left block lists the ship loading sequence and whose right block shows the corresponding dockyard container status. Cell fill styles are defined in the legend box.}}
    \label{fig:cont_plan}
\end{figure}

\subsubsection{Number of Rehandles in the Dockyard}
A rehandle in the dockyard occurs when the target container for loading is not located at the top of its assigned stack. As described in Assumption~(vi) of Section~\ref{sec:assumption}, we adopt a \emph{nearest-lowest stack strategy} to manage the placement of displaced containers. This strategy combines the advantages of both the \emph{nearest stack} and the \emph{lowest stack} approaches. While the lowest stack strategy theoretically minimizes rehandles, it is often impractical in real-world operations due to yard layout constraints and crane availability. Hence, our integrated approach prioritizes operational feasibility while still aiming to reduce rehandle occurrences.

{
\paragraph{Rehandle Count Calculation Logic}
The number of rehandles is computed as follows (see Figure{~\ref{fig:rehandle_diagram}} for a step-by-step illustration). For each loading demand event, i.e., each container that the quay crane requests from the dockyard in the order dictated by the unloading sequence $A_1$, the algorithm checks whether that container is currently at the top of its assigned yard stack. If so, the container is retrieved with no rehandles. If it is not (because one or more other containers are sitting above it), each obstructing container must be relocated to another stack according to the nearest-lowest stack rule before the target container can be accessed. Each such relocation counts as one rehandle. The relocated container is then placed on the nearest available stack with the currently lowest height, and the yard state is updated. This process repeats for every container retrieval, and the total rehandle count $R$ for a given loading sequence and dockyard plan is the sum of all individual relocation events across the entire loading operation. This total rehandle count, together with the number of single and dual cycles, enters the objective function $T = \alpha w_s + \beta w_d + \gamma R$ (Equation~1).}

\begin{figure}[!ht]
    \centering
\tikzset{
    rcont/.style={rectangle, minimum width=16mm, minimum height=9mm,
                  text centered, draw=black!70, very thick, font=\small\bfseries},
    rtarget/.style={rcont, fill=green!30, draw=green!60!black},
    rblock/.style={rcont,  fill=orange!40, draw=orange!70!black},
    rmoved/.style={rcont,  fill=orange!20, draw=orange!50!black, dashed},
    rother/.style={rcont,  fill=gray!18,   draw=gray!55},
    rcrn/.style={very thick, ->, >=stealth, color=blue!65!black},
    rstep/.style={font=\small\bfseries, align=center}
}

\begin{center}
\begin{tikzpicture}[x=1cm, y=1cm]


\draw[->, thick] (-0.5, 1.7) -- (-0.5, 4.3);
\node[font=\scriptsize, rotate=90, anchor=south] at (-0.85, 3.0) {tier increases upward};

\node[rstep] at (1.0, 5.7) {Step 1\\Initial state};
\node[rblock]  at (1.0, 3.9) {\texttt{B1}};  
\node[rblock]  at (1.0, 2.8) {\texttt{B2}};  
\node[rtarget] at (1.0, 1.7) {\texttt{T}};   
\node[font=\scriptsize] at (2.35, 3.9) {Tier 3};
\node[font=\scriptsize] at (2.35, 2.8) {Tier 2};
\node[font=\scriptsize] at (2.35, 1.7) {Tier 1};
\draw[very thick] (0.15, 1.15) -- (1.85, 1.15);
\node[font=\footnotesize\itshape] at (1.0, 0.8) {Stack $k$};
\draw[rcrn] (1.0, 5.2) -- (1.0, 4.5);
\node[font=\scriptsize, align=center] at (1.0, 5.05) {};
\node[font=\scriptsize, align=center, anchor=west] at (1.55, 4.85) {crane accesses top};

\draw[rcrn] (3.0, 3.0) -- (4.1, 3.0) node[midway, above, font=\scriptsize] {move \texttt{B1}};

\node[rstep] at (5.6, 5.7) {Step 2\\Relocate \texttt{B1} ($+1$)};
\node[rblock]  at (5.2, 2.8) {\texttt{B2}};
\node[rtarget] at (5.2, 1.7) {\texttt{T}};
\draw[very thick] (4.35, 1.15) -- (6.05, 1.15);
\node[font=\footnotesize\itshape] at (5.2, 0.8) {Stack $k$};
\node[rother]  at (6.9, 1.7) {\texttt{X}};
\node[rmoved]  at (6.9, 2.8) {\texttt{B1}};
\draw[very thick] (6.05, 1.15) -- (7.75, 1.15);
\node[font=\footnotesize\itshape] at (6.9, 0.8) {Stack $k{+}1$};

\draw[rcrn] (8.0, 3.0) -- (9.1, 3.0) node[midway, above, font=\scriptsize] {move \texttt{B2}};

\node[rstep] at (10.6, 5.7) {Step 3\\Relocate \texttt{B2} ($+1$)};
\node[rtarget] at (10.2, 1.7) {\texttt{T}};
\draw[very thick] (9.35, 1.15) -- (11.05, 1.15);
\node[font=\footnotesize\itshape] at (10.2, 0.8) {Stack $k$};
\node[rother]  at (11.9, 1.7) {\texttt{X}};
\node[rmoved]  at (11.9, 2.8) {\texttt{B1}};
\node[rmoved]  at (11.9, 3.9) {\texttt{B2}};
\draw[very thick] (11.05, 1.15) -- (12.75, 1.15);
\node[font=\footnotesize\itshape] at (11.9, 0.8) {Stack $k{+}1$};
\draw[rcrn] (10.2, 3.4) -- (10.2, 2.3);
\node[font=\scriptsize\itshape, align=center] at (10.2, 4.0) {\texttt{T} now on top\\retrieve: 0 extra};

\draw[black, thick, rounded corners=2pt] (0.0, -0.4) rectangle (12.8, -1.6);
\node[rtarget, minimum width=9mm, minimum height=6mm] at (0.7, -1.0) {};
\node[right, font=\footnotesize] at (1.2, -1.0) {Target \texttt{T} (Tier 1)};
\node[rblock, minimum width=9mm, minimum height=6mm]  at (4.1, -1.0) {};
\node[right, font=\footnotesize] at (4.6, -1.0) {Blocking (above \texttt{T})};
\node[rmoved, minimum width=9mm, minimum height=6mm]  at (7.7, -1.0) {};
\node[right, font=\footnotesize] at (8.2, -1.0) {Relocated};
\node[rother, minimum width=9mm, minimum height=6mm]  at (10.4, -1.0) {};
\node[right, font=\footnotesize] at (10.9, -1.0) {Other};

\node[font=\small\bfseries] at (6.4, -2.2)
    {Total rehandles to retrieve \texttt{T}: \textbf{2} (one per blocking container directly above \texttt{T})};

\end{tikzpicture}
\end{center}
    \caption{\rh{Conceptual illustration of the dockyard rehandle calculation. In every column, the highest drawn cell is the top of the physical stack, which is the only end the crane can access; tier numbers increase upward, with Tier~1 at the bottom. The target container \texttt{T} sits at Tier~1 of Stack~$k$, directly beneath two blocking containers, \texttt{B1} (Tier~3, top) and \texttt{B2} (Tier~2). Retrieval proceeds from the top down: \texttt{B1} is relocated first and \texttt{B2} second, each incurring one rehandle, after which \texttt{T} is exposed and retrieved, giving two rehandles in total.}}
    \label{fig:rehandle_diagram}
\end{figure}

For illustration, consider the example in Figure{~\ref{fig:cont_plan}}(b). Given the specific loading sequence and the application of the nearest-lowest stack strategy, the total number of rehandles required is three.

\section{Methodology}
\label{method}

\subsection{Algorithmic Framework}
The QCDC problem is formulated as a two-machine flow-shop scheduling problem, with unloading and loading operations as interdependent tasks. To enable tractable optimization, the following assumptions are adopted:

\subsubsection{Assumptions}\label{sec:assumption}
\begin{enumerate}[label=(\roman*)]
\item Containers designated for loading are assumed to be readily available at the dockside upon request.
\item Unloaded containers are immediately transferred from the quay area to their designated yard locations without delay.
\item Rehandle operations aboard the vessel are considered for both unloading and loading processes.
\item Containers temporarily rehandled during shipboard operations are returned to their original stack positions.
\item Shipboard rehandles are assumed to involve container movement between the vessel and the apron. In practice, some rehandles may be internal to the ship, but this simplification does not alter the optimization structure.
\item Dockyard rehandles follow the nearest-lowest stack strategy, in which a displaced container is placed in the nearest available stack with the lowest height.
\item The ship's turnaround time is used as a proxy for QC operational efficiency. It is approximated by the total number of SCs ($w_s$) and DCs ($w_d$) required for loading and unloading operations.
\item Operations are conducted row-wise: unloading and loading are completed for one row before the crane is repositioned longitudinally to the next. Due to practical limitations in QC lateral movement, dual cycling is restricted to containers within the same row.
\item No delays are considered from external factors such as inbound vehicle unavailability or crane conflicts.
\end{enumerate}

\subsubsection{Notation}
\rh{For readability, Table~\ref{tab:abbrev} first collects the abbreviations used throughout this paper together with their full forms.} The mathematical formulation of the QCDC-DR-GA problem involves a set of indices, parameters, and decision variables that define the spatial configuration of the container yard and the operational sequencing of tasks. Table~\ref{tab:notation} summarizes the notational elements used throughout the model, categorized by their roles in indexing, parameterizing, and decision-making within the optimization framework.

\rh{%
\begin{table}[!ht]
    \centering
    \caption{List of abbreviations used in this paper and their full forms.}
    \label{tab:abbrev}
    \begin{tabular}{l l}
        \toprule[1.5pt]
        \textbf{Abbreviation} & \textbf{Full form} \\
        \midrule[1pt]
        QC      & Quay Crane \\
        QCSC    & Quay Crane Single Cycling \\
        QCDC    & Quay Crane Dual Cycling \\
        QCDCS   & Quay Crane Dual-Cycling Scheduling \\
        SC      & Single Cycle \\
        DC      & Dual Cycle \\
        QCDC-DR-GA & Quay Crane Dual Cycle--Dockyard Rehandle Genetic Algorithm (proposed) \\
        GA      & Genetic Algorithm \\
        PGA     & Probabilistic Genetic Algorithm \\
        EA      & Evolutionary Algorithm \\
        ILS     & Iterated Local Search \\
        GA-ILSRS & GA with Iterated Local Search for Rehandle/Stacking (baseline) \\
        DRL     & Deep Reinforcement Learning \\
        DDQN    & Double Deep Q-Network \\
        MIP     & Mixed-Integer Programming \\
        QCSAP   & Quay Crane Scheduling and Assignment Problem \\
        AGV     & Automated Guided Vehicle \\
        TOS     & Terminal Operating System \\
        TEU     & Twenty-foot Equivalent Unit \\
        ULCV    & Ultra-Large Container Vessel \\
        \bottomrule[1.5pt]
    \end{tabular}
\end{table}
}

\begin{table}[!ht]
    \caption{Notational summary used in the QCDC-DR-GA mathematical formulation, including indices, parameters, and decision variables.}
    \centering
    \begin{tabular}{lc>{\raggedright\arraybackslash}p{0.6\linewidth}}
    \toprule[1.5pt]
    \textbf{Category} & \textbf{Symbol} & \textbf{Description} \\
    \midrule[1pt]
    \multirow{4}{*}{\textbf{Indices}} 
    & $m$ & Bay index for containers in the yard \\
    & $n$ & Stack index within a bay \\
    & $o$ & Tier index within a stack \\
    & $h_{mn}$ & Height of the yard at bay $m$ and stack $n$ \\
    \midrule[.8pt]
    \multirow{15}{*}{\textbf{Parameters}} 
    & $S$ & Set of stacks in a yard row \\
    & $U_{c}$ & Number of containers to unload from stack $c \in S$ \\
    & $L_{c}$ & Number of containers to load into stack $c \in S$ \\
    & $TU_{c}$ & Completion time of unloading stack $c \in S$ \\
    & $TL_{c}$ & Completion time of loading stack $c \in S$ \\
    & $T$ & Total completion time for both unloading and loading \\
    & $R$ & Total number of rehandles in a row during dockyard operations \\
    & $w_s$ & Total number of SCs \\
    & $w_d$ & Total number of DCs \\
    & $w$ & Total number of crane cycles, defined as $w = w_s + w_d$ \\
    & $\alpha$ & Time (seconds) required per Single Cycle crane operation \\
    & $\beta$ & Time (seconds) required per Dual Cycle crane operation \\
    & $\gamma$ & Time (seconds) required per dockyard rehandle operation \\
    & $\mu$ & A sufficiently large constant (big-M) used in sequencing constraints \\
    & $H_{mn}$ & Maximum permitted stack height at yard bay $m$, stack $n$ \\
    \midrule[0.8pt]
    \multirow{3}{*}{\textbf{Decision Variables}} 
    & $X_{ij}$ & Binary variable: 1 if unloading job $j \in S$ follows $i \in S$, 0 otherwise \\
    & $Y_{ij}$ & Binary variable: 1 if loading job $j \in S$ follows $i \in S$, 0 otherwise \\
    & $x_{rmno}$ & Binary variable: 1 if container $(m, n, o)$ is loaded onto ship bay $r$, 0 otherwise \\
    \bottomrule[1.5pt]
    \end{tabular}
    \label{tab:notation}
\end{table}

\subsubsection{Problem Setup}
The objective of the scheduling problem is to minimize the maximum completion time across all unloading and loading tasks, subject to operational constraints. The total completion time, denoted as $T$, is modeled as a weighted combination of the number of SCs ($w_s$), DCs ($w_d$), and dockyard rehandles ($R$), as follows:
\begin{equation}
    T = \alpha w_s + \beta w_d + \gamma R
\end{equation}
The corresponding optimization objective is:
\begin{equation}
    \label{eq:obj_func}
    \min T_{\max}
\end{equation}
subject to:
\begin{align}
    TL_{c} - TU_{c} &\geq \alpha L_{c} \quad &&\forall c \in S \label{eq:con1}\\
    TU_{i} - TU_{j} + \mu X_{ij} &\geq \alpha U_{i} \quad &&\forall i, j \in S \label{eq:con2}\\
    TU_{j} - TU_{i} + \mu (1 - X_{ij}) &\geq \alpha U_{j} \quad &&\forall i, j \in S \label{eq:con3}\\
    TL_{i} - TL_{j} + \mu Y_{ij} &\geq \alpha L_{i} \quad &&\forall i, j \in S \label{eq:con4}\\
    TL_{j} - TL_{i} + \mu (1 - Y_{ij}) &\geq \alpha L_{j} \quad &&\forall i, j \in S \label{eq:con5}\\
    TU_{c} &\geq \alpha U_{c} \quad &&\forall c \in S \label{eq:con6}\\
    h_{mn} &\leq H_{mn} \quad &&\forall m,n \label{eq:con7}\\
    X_{ij} &\in \{0,1\} \quad &&\forall i, j \in S \label{eq:con8}\\
    Y_{ij} &\in \{0,1\} \quad &&\forall i, j \in S \label{eq:con9}\\
    x_{rmno} &\in \{0,1\} \quad &&\forall r, m, n, o \label{eq:con10}
\end{align}

Constraints~\eqref{eq:con1}--\eqref{eq:con6} govern the sequencing and feasibility of unloading and loading tasks. Constraint~\eqref{eq:con1} ensures that loading begins only after unloading of relevant stacks is completed. Constraints~\eqref{eq:con2}--\eqref{eq:con3} enforce a valid unloading sequence, ensuring a sufficient time gap between dependent tasks. Similarly, Constraints~\eqref{eq:con4}--\eqref{eq:con5} apply the same logic to loading operations. Constraint~\eqref{eq:con6} guarantees that the time allocated for unloading each stack is sufficient. Constraint~\eqref{eq:con7} enforces yard capacity limits by restricting stack height. Constraints~\eqref{eq:con8}--\eqref{eq:con10} impose binary conditions on sequencing and placement decisions.

\subsection{Proposed QCDC-DR-GA}

\subsubsection{Computational Complexity}
\label{complexity}
The complexity of the proposed problem arises from the combinatorial nature of scheduling and container placement. Let $S$ be the number of stacks involved in ship operations. Determining the optimal unloading sequence involves evaluating all possible permutations of these stacks, leading to $S!$ candidate sequences. For each sequence, the algorithm seeks to maximize the number of DCs, further increasing computational overhead. Additionally, let $N$ represent the number of containers staged in the dockyard. The total number of possible container arrangements is $N!$, as each configuration may influence the number of rehandles. Therefore, solving both components jointly, unloading sequence optimization and dockyard rehandle minimization, yields a combined search space of size $S! \times N!$. This factorial increase in complexity makes the problem intractable for exact methods at scale, rendering it NP-hard. Consequently, heuristic and metaheuristic approaches, such as the proposed QCDC-DR-GA algorithm, are employed to obtain high-quality solutions within a reasonable computational time.

\begin{algorithm}[!ht]
    \caption{QCDC-DR-GA}
    \label{alg:genetic_algorithm}
    \KwIn{The Unloading plan and loading plan of a particular row of a ship}
    \KwOut{
    \begin{enumerate}[label=(\roman*)]
        \item Unloading sequence of the stacks of the input row
        \item Dockyard plan
    \end{enumerate}
    }
    
    Input the unloading plan and loading plan.\\
    Initialize the first generation of population with an $unloading\_sequence$ and $dockyard\_plan$\\
    \For{i $\in$ consecutive iterations}{
        \textbf{Crossover:} Use \textit{Two-Point crossover} for 1D chromosome (unloading sequence) and \textit{Two-Dimensional Substring Crossover} for 2D chromosome (dockyard plan).\\
        \textbf{Mutation:} Use \textit{Swap-Mutation} for 1D chromosome and \textit{Two-Dimensional Two-Point Swaping Mutaiton} for 2D chromosome.\\
        \textbf{Calculate Fitness:} Rearrange the Unloading-Sequences and Dockyard-Plan according to cost.\\
        \textbf{Selection:} Do \textit{Roulette Wheel} selection.\\
        \textbf{New Population:} Generate the new generation as initial population.
    }
    Output final \textit{Unloading Sequence} and \textit{Dockyard Plan}
\end{algorithm}

\subsubsection{Solution Approach}
\label{solution_approach}
In computer science and operations research, GAs are well-established metaheuristic methods inspired by the principles of natural selection. As a subset of evolutionary algorithms (EAs), GAs are capable of producing high-quality solutions for complex optimization problems through iterative refinement of a population of candidate solutions. In this study, we design a hybrid GA that operates on a mixed chromosome structure consisting of: (i) a one-dimensional unloading sequence, and (ii) a two-dimensional dockyard container arrangement. The algorithm employs tailored crossover and mutation operators for each component to ensure feasible offspring and maintain search diversity. A key challenge in our QCDC-DR-GA framework is fitness evaluation: accurately calculating the cost of each candidate solution. This cost depends on both the unloading sequence and the resulting dockyard rehandling effort. Integrating these two components effectively is crucial for evaluating the total turnaround time. The overall procedure of the proposed algorithm is summarized in Algorithm~\ref{alg:genetic_algorithm}.

{
\subsubsection{Initial Population Generation}
\label{sec:initial_pop}
The initial population, denoted by $P$, consists of chromosomes encoding both unloading sequences and dockyard stacking plans. Each chromosome $c_i$ comprises two components: the one-dimensional unloading sequence $c_{i(us)}$ and the two-dimensional dockyard plan $c_{i(dp)}$. \rh{These two components correspond directly to the physical layouts of Figure~\ref{fig:row_tier_bay}. The unloading-sequence vector $A_1$ encodes the order in which the stacks of a single row are processed by the quay crane; each entry of $A_1$ is therefore a stack index $1,\dots,N$ corresponding to the side-by-side stacks visible in the front view of Figure~\ref{fig:row_tier_bay}(b). The dockyard-plan matrix $A_2$ encodes the yard arrangement of the containers that will be loaded: each row of $A_2$ is a dockyard stack, and each column is a tier position within that stack, with Tier~1 at the bottom. A worked example mapping $A_1$ and $A_2$ to these physical units is given below.}

\begin{table}[H]
\begin{tabular}{ll}
    $P$ & : Set of chromosomes in the population \\
    $n$ & : Number of chromosomes in $P$ \\
    $c_i$ & : The $i$-th chromosome, where $1 \leq i \leq n$ \\
    $c_{i(us)}$ & : Unloading sequence component of chromosome $c_i$ \\
    $c_{i(dp)}$ & : Dockyard plan component of chromosome $c_i$ \\
\end{tabular}
\end{table}

An example of a candidate solution is illustrated below:

\[
A_1 : \left[ \begin{array}{cccccccccc}
1 & 2 & 3 & 4 & 5 & 6 & 7 & 8 & 9 & 10
\end{array} \right]
\quad \text{(Unloading Sequence)}
\]

\[
A_2 : \left[ \begin{array}{cccc}
3A & 3B & 1C & 1D \\
1A & 2B &     &     \\
2A & 1E & 1B &     \\
3C & 2C & 3E & 3D \\
2D & 2E & 4A & 4D \\
4B & 4C & 3C &     \\
4E &     &     &    
\end{array} \right]
\quad \text{(Dockyard Plan)}
\]

In matrix $A_2$, each element represents a container, where the alphanumeric label encodes the intended position on the ship \rh{using the global \texttt{[stack][tier]} convention defined in Section~\ref{problem}}. For example, the value `3A' indicates that the container is scheduled to be loaded into \rh{ship stack~3 at tier~A, i.e., Tier~1, the bottom of that stack}. \rh{Correspondingly, the entries of $A_1$ are the canonical numeric stack indices $1,2,\dots,N$: the value at position $k$ of $A_1$ is the index of the ship stack processed $k$-th by the quay crane. Thus, the integers in $A_1$ and the leading digit of each container label in $A_2$ refer to the same set of ship-stack indices, so the two chromosome components share a single, consistent stack-numbering scheme.} Thus, $A_1$ and $A_2$ correspond to $c_{i(us)}$ and $c_{i(dp)}$, respectively, and together define a complete chromosome in the population.
}

\subsubsection{Design Rationale: Joint Optimization of $A_1$ and $A_2$}
\label{sec:joint_design}
A central design choice in QCDC-DR-GA is that the dockyard plan $A_2$ is not fixed before running the GA; rather, it is treated as an active decision variable that evolves jointly with the unloading sequence $A_1$ throughout the entire optimization loop. This integration is intentional and theoretically motivated. The reason is as follows. In scheduled liner services, a vessel's departure from its previous port generates a complete loading manifest well before arrival at the next port. This manifest specifies exactly which containers must be loaded, their intended ship positions, and the approximate arrival sequence of the containers at the dockside. This window, typically 12–36 hours, provides a realistic opportunity for port planners to reconfigure the dockyard arrangement $A_2$ in preparation for the vessel's arrival, a process referred to as \emph{pre-arrival preparation}. By embedding $A_2$ optimization within the GA loop, QCDC-DR-GA converts this available pre-arrival window into a quantifiable operational efficiency gain. Computationally, this integration works as follows. Each chromosome $c_i = (c_{i(us)}, c_{i(dp)})$ encodes both the unloading sequence and the dockyard plan as a single genetic unit. During fitness evaluation (Algorithm{~\ref{alg:cost_1}}), the cost function simultaneously simulates the dual-cycling operation using the proposed $A_1$ and counts the dockyard rehandles implied by $A_2$ given the same unloading order. This coupled evaluation means that the genetic operators, crossover and mutation, apply selection pressure on both components simultaneously, allowing the algorithm to discover configurations where the unloading order and the dockyard layout are mutually reinforcing rather than conflicting. The cost of this integration is an increase in chromosome dimensionality and therefore evaluation complexity; however, as shown in Section{~\ref{complexity}}, the combined search space of $S! \times N!$ is handled efficiently by the metaheuristic framework within the tested experimental ranges.

\subsubsection{Crossover}
To generate offspring for the next generation, two parent solutions are selected from the current population, and their genetic material is recombined using crossover operators. Since each solution consists of both one-dimensional and two-dimensional chromosome components, we employ separate crossover strategies tailored to each part.

\paragraph{One-Dimensional Crossover}
For the 1D chromosome, we apply the \textit{Two-Point Crossover} method, a special case of the more general \textit{N-Point Crossover}. Two crossover points are randomly selected along the parent chromosomes, and the segments between these points are exchanged to form the offspring. To ensure solution validity, duplicate genes are removed, and any missing genes are appended to the end of the chromosome in the order they appear \rh{; this duplicate-removal and missing-gene-repair step guarantees that every offspring remains a valid permutation of the stack indices} (see Figure~\ref{1D_cross}).

\begin{figure}[!htbp]
    \centering

\tikzstyle{box1} = [rectangle, minimum size = 8mm, text centered, draw=green!=60, fill=green!8, very thick]
\tikzstyle{box2} = [rectangle, minimum size = 8mm, text centered, draw=blue!=60, fill=blue!8, very thick]
\tikzstyle{box3} = [rectangle, minimum size = 8mm, text centered, draw=green!=60, fill=red!20, very thick]
\tikzstyle{box4} = [rectangle, minimum size = 8mm, text centered, draw=blue!=60, fill=red!20, very thick]
\tikzstyle{box5} = [rectangle, minimum size = 8mm, text centered, draw=green!=100, fill=green!50, very thick]
\tikzstyle{box6} = [rectangle, minimum size = 8mm, text centered, draw=blue!=100, fill=blue!50, very thick]
\tikzstyle{arrow} = [line width=0.5mm, -{Stealth[length=4mm, open, round]}]

\begin{center}
\resizebox{0.7\linewidth}{!}{
    \begin{tikzpicture}
        \foreach \i in {1,...,12} {
            \node (\i) [box1] at (\i-1,-1) {\i};
        }
        \node (p1) [left of = 1, xshift = -1cm] {Parent 1};

        \begin{scope}[yshift=-23mm]
            \foreach \i in {12,...,1} {
                \node (\i') [box2] at (12 - \i, 0) {\i};
            }
        \end{scope}
        \node (p2) [left of = 12', xshift = -1cm] {Parent 2};

        \begin{scope}[yshift=-40mm]
            \foreach \i in {1,...,3} {
                \node (\i'') [box1] at (\i - 1, 0) {\i};
            }
            \node (9'') [box4] at (3, 0) {9};
            \foreach \i in {8,...,5} {
                \node (\i'') [box2] at (12 -\i, 0) {\i};
            }
            \foreach \i in {9,...,12} {
                \node (\i'') [box1] at (\i - 1, 0) {\i};
            }
            \node (o1) [left of = 1'', xshift = -1cm] {Offspring 1};
        \end{scope}

        \begin{scope}[yshift=-53mm]
            \foreach \i in {12,...,10} {
                \node (\i''') [box2] at (12 - \i, 0) {\i};
            }
            \foreach \i in {4,...,8} {
                \node (\i''') [box1] at (\i - 1, 0) {\i};
            }
            \node (4''') [box3] at (3, 0) {4}; 
            \foreach \i in {4,...,1} {
                \node (\i''') [box2] at (12 -\i, 0) {\i};
            }
            \node (o2) [left of = 12''', xshift = -1cm] {Offspring 2};
        \end{scope}

        \draw[dashed, line width = 1.5pt] (2.5, 0) -- (2.5, -6.2);
        \draw[dashed, line width = 1.5pt] (7.5, 0) -- (7.5, -6.2);

        \begin{scope}[yshift=-70mm]
            \foreach \i in {1,...,3} {
                \node (\i'') [box1] at (\i - 1, 0) {\i};
            }
            \foreach \i in {8,...,5} {
                \node (\i'') [box2] at (11 -\i, 0) {\i};
            }
            \foreach \i in {9,...,12} {
                \node (\i'') [box1] at (\i - 2, 0) {\i};
            }
            \node (4'') [box5] at (11, 0) {4};
            \node (o11) [left of = 1'', xshift = -1cm] {Offspring 1};
        \end{scope}

        \begin{scope}[yshift=-83mm]
            \foreach \i in {12,...,10} {
                \node (\i''') [box2] at (12 - \i, 0) {\i};
            }
            \foreach \i in {5,...,8} {
                \node (\i''') [box1] at (\i - 2, 0) {\i};
            }
            \foreach \i in {4,...,1} {
                \node (\i''') [box2] at (11 -\i, 0) {\i};
            }
            \node (9''') [box6] at (11, 0) {9}; 
            \node (o22) [left of = 12''', xshift = -1cm] {Offspring 2};
        \end{scope}

        \draw[arrow] (5.5, -2.8) -- (5.5,- 3.5);
        \draw[arrow] (5.5, -5.8) -- (5.5,- 6.5);

        \draw[-stealth, bend right=30, very thick] (-3, -4.2) to (-3, -6.7);
        \draw[-stealth, bend right=30, very thick] (-3, -5.5) to (-3, -7.9);
    \end{tikzpicture}
    }
\end{center}


    \caption{1D \textit{Two-Point Crossover} technique. \rh{Two cut-points (the dashed vertical lines) are randomly selected, and the segment between them is exchanged between the two parents: Parent~1's genes are drawn on a light-green fill, and Parent~2's on a light-blue fill. After the swap, a gene that appears twice in an offspring is marked on a red fill (gene~9 in Offspring~1 and gene~4 in Offspring~2) and is removed; the missing gene is then appended, shown on its original-parent color (gene~4 on green in Offspring~1, gene~9 on blue in Offspring~2).}}
    \label{1D_cross}
\end{figure}

\paragraph{Two-Dimensional Crossover}
For the two-dimensional solution representation, we adopt a \textit{2D Substring Crossover} strategy inspired by the aircraft gate assignment problem \citep{tsai2015two}. This operator performs both row-wise and column-wise genetic exchanges to generate diverse offspring while preserving structural feasibility. In the \textit{row-wise crossover}, two random rows are selected as boundaries, and the corresponding submatrices between these rows are exchanged between the parent solutions. Subsequently, a \textit{column-wise operation} is applied to the selected rows using the classical \textit{Two-Point Crossover} technique, commonly used for one-dimensional encodings. To maintain solution validity, duplicate entries are removed from the resulting offspring, and any missing elements are appended in a post-processing step. \rh{This repair step ensures that every offspring dockyard plan contains each container exactly once and respects the stack-height limit.} This ensures that all container assignments remain valid and no information is lost during recombination. An illustrative example of the 2D crossover mechanism is shown in Figure~\ref{2D_cross}.

\begin{figure}[!htbp]
    \centering
\tikzstyle{box1} = [rectangle, minimum size = 8mm, text centered, draw=green!=60, fill=green!5, very thick]
\tikzstyle{box2} = [rectangle, minimum size = 8mm, text centered, draw=blue!=60, fill=blue!5, very thick]
\tikzstyle{box3} = [rectangle, minimum size = 8mm, text centered, draw=green!=60, fill=red!20, very thick]
\tikzstyle{box4} = [rectangle, minimum size = 8mm, text centered, draw=blue!=60, fill=red!20, very thick]
\tikzstyle{box5} = [rectangle, minimum size = 8mm, text centered, draw=green!=100, fill=green!50, very thick]
\tikzstyle{box6} = [rectangle, minimum size = 8mm, text centered, draw=blue!=100, fill=blue!50, very thick]

\tikzstyle{arrow} = [very thick,->,>=angle 60]

\newcommand{\pari}{{4C, 2B}, {2C, 4A, 3A}, {1A, 1B, 1C}, {2A, 4B}, {2D, 3B}}
\newcommand{\parii}{{3A, 4B, 3B}, {1A, 2D, 4C}, {2B, 4A}, {1B}, {2A, 2C, 1C}}

\begin{center}
\resizebox{.5\linewidth}{!}{
    \begin{tikzpicture}
        \setcounter{c}{1}
        \node (p1) at (0.5, 0) {Parent 1};
        \node (p2) at (8.5, 0) {Parent 2};
        \foreach \row in \pari {
            \setcounter{r}{0}
            \foreach \i in \row {
                \node (\i) [box1] at (\value{r}, -\value{c}) {\i};
                \stepcounter{r}
            }
            \stepcounter{c}
        }

        \setcounter{c}{1}
        \foreach \row in \parii {
            \setcounter{r}{8}
            \foreach \i in \row {
                \node (\i) [box2] at (\value{r}, -\value{c}) {\i};
                \stepcounter{r}
            }
            \stepcounter{c}
        }
        \draw[dashed, line width = 1.5pt] (-1, -1.5) -- (11, -1.5);
        \draw[dashed, line width = 1.5pt] (-1, -3.5) -- (11, -3.5);

        \node (o1) at (1.3, -\value{c}) {Offspring 1 (row-swap)};
        \node (o2) at (9.3, -\value{c}) {Offspring 2 (row-swap)};

        \stepcounter{c}
        \newcounter{cnt}
        \setcounter{cnt}{0}
        \foreach \row in \pari {
            \setcounter{r}{0}
            \ifnum \value{cnt} < 1
                \foreach \i in \row {
                    \node (\i) [box1] at (\value{r}, -\value{c}) {\i};
                    \stepcounter{r}
                }
                \stepcounter{c}
            \fi
            \stepcounter{cnt}
        }
        \setcounter{cnt}{0}
        \foreach \row in \parii {
            \setcounter{r}{0}
            \ifnum \value{cnt} > 0
                \ifnum \value{cnt} < 3
                    \foreach \i in \row {
                        \node (\i) [box2] at (\value{r}, -\value{c}) {\i};
                        \stepcounter{r}
                    }
                    \stepcounter{c}
                \fi
            \fi
            \stepcounter{cnt}
        }
        \setcounter{cnt}{0}
        \foreach \row in \pari {
            \setcounter{r}{0}
            \ifnum \value{cnt} > 2
                \foreach \i in \row {
                    \node (\i) [box1] at (\value{r}, -\value{c}) {\i};
                    \stepcounter{r}
                }
                \stepcounter{c}
            \fi
            \stepcounter{cnt}
        }

        \setcounter{c}{7}
        \setcounter{cnt}{0}
        \foreach \row in \parii {
            \setcounter{r}{8}
            \ifnum \value{cnt} < 1
                \foreach \i in \row {
                    \node (\i) [box2] at (\value{r}, -\value{c}) {\i};
                    \stepcounter{r}
                }
                \stepcounter{c}
            \fi
            \stepcounter{cnt}
        }
        \setcounter{cnt}{0}
        \foreach \row in \pari {
            \setcounter{r}{8}
            \ifnum \value{cnt} > 0
                \ifnum \value{cnt} < 3
                    \foreach \i in \row {
                        \node (\i) [box1] at (\value{r}, -\value{c}) {\i};
                        \stepcounter{r}
                    }
                    \stepcounter{c}
                \fi
            \fi
            \stepcounter{cnt}
        }
        \setcounter{cnt}{0}
        \foreach \row in \parii {
            \setcounter{r}{8}
            \ifnum \value{cnt} > 2
                \foreach \i in \row {
                    \node (\i) [box2] at (\value{r}, -\value{c}) {\i};
                    \stepcounter{r}
                }
                \stepcounter{c}
            \fi
            \stepcounter{cnt}
        }

        \draw[black, line width = 1.5pt, dashed] (-0.5, -8.5) rectangle (0.5, -7.5);
        \draw[black, line width = 1.5pt, dashed] (7.5, -8.5) rectangle (8.5, -7.5);
        \draw[black, line width = 1.5pt, dashed] (0.5, -9.5) rectangle (2.5, -8.5);
        \draw[black, line width = 1.5pt, dashed] (8.5, -9.5) rectangle (10.5, -8.5);

        \node (o1) at (1.5, -\value{c}) {Offspring 1 (column-swap)};
        \node (o2) at (9.5, -\value{c}) {Offspring 2 (column-swap)};

        \stepcounter{c}
        \setcounter{cnt}{0}
        \foreach \row in \pari {
            \setcounter{r}{0}
            \ifnum \value{cnt} < 1
                \foreach \i in \row {
                    \node (\i) [box1] at (\value{r}, -\value{c}) {\i};
                    \stepcounter{r}
                }
                \stepcounter{c}
            \fi
            \stepcounter{cnt}
        }
        \setcounter{r}{0}
        \node (2c) [box1] at (\value{r}, -\value{c}) {2C};
        \stepcounter{r}
        \node (2d) [box2] at (\value{r}, -\value{c}) {2D};
        \stepcounter{r}
        \node (4c) [box2] at (\value{r}, -\value{c}) {4C};
        \stepcounter{c}
        \setcounter{r}{0}
        \node (2b) [box4] at (\value{r}, -\value{c}) {2B};
        \stepcounter{r}
        \node (1b) [box1] at (\value{r}, -\value{c}) {1B};
        \stepcounter{r}
        \node (1c) [box1] at (\value{r}, -\value{c}) {1C};
        
        \stepcounter{c}
        \setcounter{cnt}{0}
        \foreach \row in \pari {
            \setcounter{r}{0}
            \ifnum \value{cnt} > 2
                \foreach \i in \row {
                    \node (\i) [box1] at (\value{r}, -\value{c}) {\i};
                    \stepcounter{r}
                }
                \stepcounter{c}
            \fi
            \stepcounter{cnt}
        }

        \setcounter{c}{13}
        \setcounter{cnt}{0}
        \foreach \row in \parii {
            \setcounter{r}{8}
            \ifnum \value{cnt} < 1
                \foreach \i in \row {
                    \node (\i) [box2] at (\value{r}, -\value{c}) {\i};
                    \stepcounter{r}
                }
                \stepcounter{c}
            \fi
            \stepcounter{cnt}
        }
        
        \setcounter{r}{8}
        \node (1c) [box4] at (\value{r}, -\value{c}) {1C};
        \stepcounter{r}
        \node (4a) [box1] at (\value{r}, -\value{c}) {4A};
        \stepcounter{r}
        \node (3a) [box3] at (\value{r}, -\value{c}) {3A};
        \stepcounter{c}
        \setcounter{r}{8}
        \node (1b) [box3] at (\value{r}, -\value{c}) {1B};
        \stepcounter{r}
        \node (4a) [box4] at (\value{r}, -\value{c}) {4A};

        \stepcounter{c}
        \setcounter{cnt}{0}
        \foreach \row in \parii {
            \setcounter{r}{8}
            \ifnum \value{cnt} > 2
                \foreach \i in \row {
                    \node (\i) [box2] at (\value{r}, -\value{c}) {\i};
                    \stepcounter{r}
                }
                \stepcounter{c}
            \fi
            \stepcounter{cnt}
        }

        \node (o1) at (1.2, -\value{c}) {Offspring 1 (repaired)};
        \node (o2) at (9.2, -\value{c}) {Offspring 2 (repaired)};
        
        \stepcounter{c}
        \setcounter{r}{0}
        \node (4c) [box1] at (\value{r}, -\value{c}) {4C};
        \stepcounter{r}
        \node (2b) [box1] at (\value{r}, -\value{c}) {2B};
        \stepcounter{r}
        \node (3a) [box5] at (\value{r}, -\value{c}) {3A};
        \stepcounter{c}
        \setcounter{r}{0}
        \node (2c) [box1] at (\value{r}, -\value{c}) {2C};
        \stepcounter{r}
        \node (2d) [box2] at (\value{r}, -\value{c}) {2D};
        \stepcounter{r}
        \node (4c) [box2] at (\value{r}, -\value{c}) {4C};
        \stepcounter{c}
        \setcounter{r}{0}
        \node (1b) [box1] at (\value{r}, -\value{c}) {1B};
        \stepcounter{r}
        \node (1c) [box1] at (\value{r}, -\value{c}) {1C};
        
        \stepcounter{c}
        \setcounter{cnt}{0}
        \foreach \row in \pari {
            \setcounter{r}{0}
            \ifnum \value{cnt} > 2
                \foreach \i in \row {
                    \node (\i) [box1] at (\value{r}, -\value{c}) {\i};
                    \stepcounter{r}
                }
                \stepcounter{c}
            \fi
            \stepcounter{cnt}
        }

        \setcounter{c}{19}
        \setcounter{cnt}{0}
        \foreach \row in \parii {
            \setcounter{r}{8}
            \ifnum \value{cnt} < 1
                \foreach \i in \row {
                    \node (\i) [box2] at (\value{r}, -\value{c}) {\i};
                    \stepcounter{r}
                }
                \stepcounter{c}
            \fi
            \stepcounter{cnt}
        }
        
        \setcounter{r}{8}
        \node (4a) [box1] at (\value{r}, -\value{c}) {4A};
        \stepcounter{r}
        \node (1a) [box6] at (\value{r}, -\value{c}) {1A};
        \stepcounter{r}
        \node (2d) [box6] at (\value{r}, -\value{c}) {2D};
        \stepcounter{c}
        \setcounter{r}{8}
        \node (4c) [box6] at (\value{r}, -\value{c}) {4C};
        \stepcounter{r}
        \node (2b) [box6] at (\value{r}, -\value{c}) {2B};

        \stepcounter{c}
        \setcounter{cnt}{0}
        \foreach \row in \parii {
            \setcounter{r}{8}
            \ifnum \value{cnt} > 2
                \foreach \i in \row {
                    \node (\i) [box2] at (\value{r}, -\value{c}) {\i};
                    \stepcounter{r}
                }
                \stepcounter{c}
            \fi
            \stepcounter{cnt}
        }

    \end{tikzpicture}
    }
\end{center}
    \caption{2D Two-Point Substring Crossover technique. \rh{Parent~1's dockyard plan is drawn on a light-green fill, and Parent~2's on a light-blue fill throughout. Two horizontal dashed lines mark the randomly chosen row boundaries defining the exchanged submatrix block, which is swapped between the parents to produce the ``row-swap'' offspring. A column-wise Two-Point Crossover is then applied within the selected rows (dashed boxes). Any container label duplicated after the exchange is flagged on a red fill (e.g.\ \texttt{2B} and \texttt{1B} in the column-swap stage) and removed, and the missing labels are re-inserted on their source-parent color (e.g.\ \texttt{3A} on green; \texttt{1A}, \texttt{2D}, \texttt{4C}, \texttt{2B} on blue).}}
    \label{2D_cross}
\end{figure}

\subsubsection{Mutation}
A mutation is a genetic change that maintains the genetic diversity of chromosomes across generations. It is similar to a biological mutation.
Again, we use two different methods for two parts of the chromosome. \\
\textit{Notation:}\\
 $P_m$ = The probability of mutation.\\
The mutation operation will not occur after every cross. The mutation method is described in Algorithm \ref{alg:mutation}. 

\begin{algorithm}[!ht]
    \caption{Mutation Algorithm}
    \label{alg:mutation}
    \KwIn{Two 1D vectors after newly crossed child}
    \KwOut{Two 2D vectors as mutated child}

    Generate a random number $R$.

    \If{$R > P_m$} {
        Do not do the mutation operation.
    }
    \Else {
        Do a mutation operation.
    }
\end{algorithm}

\paragraph{One-Dimensional Mutation} We use the \textit{Swap Mutation} method for the 1D chromosome segment, interchanging two selected genes after crossover (see Figure~\ref{1D_mutation}).

\begin{figure*}[!ht]
    \centering
\tikzstyle{box1} = [rectangle, minimum size = 8mm, text centered, draw=green!=60, fill=green!8, very thick]
\tikzstyle{box5} = [rectangle, minimum size = 8mm, text centered, draw=green!=100, fill=green!50, very thick]

\tikzstyle{arrow} = [very thick,->,>=angle 60]

\begin{tikzpicture}
    \foreach \i in {1,...,12} {
        \node (\i) [box1] at (\i-1, -1) {\i};
    }
    \draw[arrow] (3, 0.5) -- (3, -0.5);
    \draw[arrow] (10, 0.5) -- (10, -0.5);
    \foreach \i in {1,...,3} {
        \node (\i) [box1] at (\i - 1, -3) {\i};
    }
    \node (11) [box5] at (3, -3) {11};
    \foreach \i in {5,...,10} {
        \node (\i) [box1] at (\i - 1, -3) {\i};
    }
    \node (4) [box5] at (10, -3) {4};
    \node (12) [box1] at (11, -3) {12};
    \draw[-stealth, bend right=50, very thick] (-0.5, -1) to (-0.5, -3);
    
    \draw[black, line width = 1.5pt, dashed] (2.5, -0.5) rectangle (3.5, -1.5);
    \draw[black, line width = 1.5pt, dashed] (9.5, -0.5) rectangle (10.5, -1.5);
\end{tikzpicture}
    \caption{Swap mutation for the 1D unloading-sequence chromosome. Two gene positions (indicated by arrows) are randomly selected and their values exchanged. Because the unloading sequence is a permutation, swapping preserves validity: no gene is duplicated or lost. The mutated positions are highlighted to distinguish them from unchanged genes.}
    \label{1D_mutation}
\end{figure*}

\paragraph{Two Dimensional Mutation} Here, we chose the \textit{2D Two-Point Swapping Mutation} method for the 2D part of our chromosome. This is also the modified version of the mutation method introduced by \cite{tsai2015two}. The method is described in Algorithm \ref{alg:2D_mutation} (also see Figure \ref{2D_mutation}).\\
\textit{Notations:}\\
 $R$: The number of rows in the 2D chromosome part.\\
 $C_{R_i}$: The number of columns in the $i^{th}$ row.

\begin{figure}[!ht]
    \centering
\newcommand{\pari}{{4C, 2B}, {2C, 4A, 3A}, {1A, 1B, 1C}, {2A, 4B}, {2D, 3B}}
\tikzstyle{box1} = [rectangle, minimum size = 8mm, text centered, draw=green!=60, fill=green!5, very thick]
\tikzstyle{arrow} = [very thick, ->, >=angle 60]
\tikzstyle{box5} = [rectangle, minimum size = 8mm, text centered, draw=green!=100, fill=green!50, very thick]

\newcommand{\rowone}{2C, 4A, 3A}
\newcommand{\rowtwo}{1A, 1B, 1C}

\begin{tikzpicture}
    \centering
    \setcounter{c}{0}
    \foreach \row in \pari {
        \setcounter{r}{0}
        \foreach \i in \row {
            \node (\i) [box1] at (\value{r}, -\value{c}) {\i};
            \stepcounter{r}
        }
        \stepcounter{c}
    }

    \draw [arrow] (-1.6, -1) -- (-0.6, -1);
    \draw [arrow] (-1.6, -2) -- (-0.6, -2);

    \draw[black, line width = 1.5pt, dashed] (-0.5, -0.5) rectangle (2.5, -1.5);
    \draw[black, line width = 1.5pt, dashed] (-0.5, -1.5) rectangle (2.5, -2.5);
    
    \draw[line width=0.5mm, -{Stealth[length=5mm, open, round]}] (3.3, -1.5) -- (6.5, -1.5);

    \setcounter{r}{8}
    \foreach \i in \rowone {
        \node (\i) [box1] at (\value{r}, -0.8) {\i};
        \stepcounter{r}
    }

    \setcounter{r}{8}
    \foreach \i in \rowtwo {
        \node (\i) [box1] at (\value{r}, -2.2) {\i};
        \stepcounter{r}
    }

    \draw [arrow] (8, 0.8) -- (8, -0.2);
    \draw [arrow] (9, -3.8) -- (9, -2.8);

    \stepcounter{c}
    \setcounter{cnt}{0}
    \foreach \row in \pari {
        \setcounter{r}{4}
        \ifnum \value{cnt} < 1
            \foreach \i in \row {
                \node (\i) [box1] at (\value{r}, -\value{c}) {\i};
                \stepcounter{r}
            }
            \stepcounter{c}
        \fi
        \stepcounter{cnt}
    }
    \setcounter{r}{4}
    \node (1b) [box5] at (\value{r}, -\value{c}) {1B};
    \stepcounter{r}
    \node (4a) [box1] at (\value{r}, -\value{c}) {4A};
    \stepcounter{r}
    \node (3a) [box1] at (\value{r}, -\value{c}) {3A};
    \stepcounter{c}
    \setcounter{r}{4}
    \node (1a) [box1] at (\value{r}, -\value{c}) {1A};
    \stepcounter{r}
    \node (2c) [box5] at (\value{r}, -\value{c}) {2C};
    \stepcounter{r}
    \node (1c) [box1] at (\value{r}, -\value{c}) {1C};
    
    \stepcounter{c}
    \setcounter{cnt}{0}
    \foreach \row in \pari {
        \setcounter{r}{4}
        \ifnum \value{cnt} > 2
            \foreach \i in \row {
                \node (\i) [box1] at (\value{r}, -\value{c}) {\i};
                \stepcounter{r}
            }
            \stepcounter{c}
        \fi
        \stepcounter{cnt}
    }

    \draw[line width=0.5mm, -{Stealth[length=5mm, open, round]}, bend right=30] (7, -2.5) to (4.5, -5);

\end{tikzpicture}
    \caption{2D two-point swapping mutation applied to the dockyard plan chromosome. Two rows ($r_1$, $r_2$) are randomly selected (indicated by horizontal arrows on the left), and one column index is randomly chosen within each selected row ($c_1$ in row $r_1$, $c_2$ in row $r_2$). The genes at positions $(r_1, c_1)$ and $(r_2, c_2)$ are exchanged (highlighted cells), introducing spatial diversity into the dockyard arrangement while keeping all other container assignments intact.}
    \label{2D_mutation}
\end{figure}

\begin{algorithm}[!ht]
  \caption{2D Mutation Algorithm}
  \label{alg:2D_mutation}
  \KwIn{A 2D vector from newly crossed children}
  \KwOut{A 2D vector as a mutated child}
  
  Randomly generate $r1$ and $r2$ to select two rows from the 2D vector, where $1 \leq r1, r2 \leq R$\\
  Generate random integers $c1$ and $c2$ to select two points from the selected rows, where $1 \leq c1 \leq Cr1$ and $1 \leq c2 \leq Cr2$\\
  Interchange the genes between the selected points of the 2D vector
\end{algorithm}

\subsubsection{Calculate Fitness}
The fittest chromosomes are selected from each generation based on their cost. The cost here is the total completion time, which is nothing but our objective function. Every time a new generation is produced, the cost is calculated and stored against every chromosome. Then the population is sorted in ascending order by cost. The population is now ready for the selection stage. The calculation method of the cost of each chromosome is explained in detail in Algorithms \ref{alg:cost_1} and \ref{alg:cost_2}.

\SetKwFunction{unloadFirstStack}{unload\_first\_stack}
\SetKwProg{fn}{Function}{:}{}
\SetKwFunction{loading}{loading\_operation}
\SetKwFunction{calRehandles}{calculate\_rehandles}

\begin{algorithm}[h!b]
    \caption{Cost function}
    \label{alg:cost_1}
    \KwIn{Loading plan, unloading plan, dockyard container arrangement, maximum dockyard container stack height}
    \KwOut{No. of single cycles, no. of double cycles, no. of dockyard rehandles}

    \fn{\unloadFirstStack{unloading plan, unloading sequence}} {
        \For{container $\in$ the dockyard stack of unloadingSequence} {
            \If{the container will not stay on the vessel} {
                unload the container\\
                no\_of\_single\_cycles += 1
            }
        }
    }

    \fn{\calRehandles{target container}} {
        Let, no\_of\_rehandles $\leftarrow 0$\\
        Let, found\_the\_container $\leftarrow false$\\
        \For{$i$ $\in$ stacks of dockyard}{
            \For{$j$ $\in$ containers of current stack} {
                \If{$j=$ target container}{
                    found\_the\_container $\leftarrow true$\\
                    \While{until containers are shifted from the top of the target container one by one}{
                        no\_of\_rehandles $+= 1$\\
                        Shift the container nearest lowest stack
                    }
                    \Return{no\_of\_rehandles}
                }
            }
        }
        \If{found\_the\_container $= false$}{
            Warning! container not found
        }
        \Return{0}
    }
    
    \fn{\loading{unloading plan, loading plan, unloading sequence}} {
        \If{the current loading stack is empty} {
            \If{the current unloading stack is empty} {
                go to the next loading stack
            }
            \Else{
                \Return{false, 0}
            }
        }
        Load the current container from dockyard\\
        \Return{true, \calRehandles{current container to be loaded at dockyard}}
    }
    Let, no\_of\_single\_cycles $\leftarrow 0$\\
    Let, no\_of\_double\_cycles $\leftarrow 0$\\
    Let, no\_of\_rehandles $\leftarrow 0$\\
    \unloadFirstStack{unloading plan, unloading sequence}\\
\end{algorithm}

\begin{algorithm}[h!tbp]
    \caption{Cost function (continued)}
    \label{alg:cost_2}
    \setcounter{AlgoLine}{31}

    \While{until all the stacks are unloaded from ship}{
        \For{container $\in$ current stack}{
            \If{the container will not stay on the vessel} {
                unload the container\\
                \If{there is any container to load \textbf{and} any stack of the ship is free for loading} {
                    flag, rehandles $\leftarrow$ \loading{unloadig plan, loading plan, unloading sequence}\\
                    no\_of\_rehandles += rehandles\\
                    no\_of\_dual\_cycles += 1
                }
            }
        }
    }

    \While{complete loading the remaining stacks}{
        flag, rehandles $\leftarrow$ \loading{unloadig plan, loading plan, unloading sequence}\\
        no\_of\_rehandles += rehandles\\
        no\_of\_single\_cycles += 1
    }
    
\end{algorithm}

\begin{figure}[!ht]
    \centering
\newcommand{\arrdata}{%
  30/8.3/C_1/667,%
  40/2.7/C_2/820,%
  90/14/C_3/559,%
  110/5.5/C_4/763,%
  140/8.3/C_5/667,%
  158/5/C_6/780,%
  187/8/C_7/704,%
  237/13/C_8/572,%
  277/11/C_9/600,%
  305/7.8/C_{10}/694,%
  320/4/C_{11}/850,%
  360/11/C_{12}/604}

\begin{tikzpicture}
  \draw[black, very thick] (0,0) circle (3cm);

  \foreach \endang/\pct/\cid/\tval in \arrdata {
    \draw[black, very thick] (0,0) -- ({\endang}:3cm);
  }

  \foreach \mid/\pct in {%
      15/8.3, 35/2.7, 65/14, 100/5.5, 125/8.3,
      149/5, 173/8, 212/13, 257/11, 291/7.8, 313/4, 340/11} {
    \node[anchor=center, font=\tiny\bfseries] at ({\mid}:2.65cm) {\pct\%};
  }

  \foreach \mid/\cid in {%
      15/$C_1$, 35/$C_2$, 65/$C_3$, 100/$C_4$,
      125/$C_5$, 149/$C_6$, 173/$C_7$, 212/$C_8$,
      257/$C_9$, 291/$C_{10}$, 313/$C_{11}$, 340/$C_{12}$} {
    \node[anchor=center, font=\tiny] at ({\mid}:1.9cm) {\cid};
  }

  \foreach \mid/\tval in {%
      15/667, 35/820, 65/559, 100/763,
      125/667, 149/780, 173/704, 212/572,
      257/600, 291/694, 313/850, 340/604} {
    \node[anchor=center, font=\tiny\itshape] at ({\mid}:1.2cm) {\tval\,min};
  }

  \fill[black] (0,0) circle (0.1cm);

  \node[font=\scriptsize\bfseries,  align=center] at (0,-3.55)
    {Outer: selection probability (\%)  $\cdot$  Middle: chromosome ID  $\cdot$  Inner: operation time (min)};
  \node[font=\scriptsize\itshape, align=center] at (0,-4.05)
    {Larger arc $\Rightarrow$ lower operation time $\Rightarrow$ higher fitness $\Rightarrow$ higher selection probability};
\end{tikzpicture}
    \caption{Illustration of the weighted roulette wheel used for parent selection. Each segment corresponds to a chromosome (individual) in the current population, labeled by its individual ID (e.g., $C_1, C_2, \ldots$). The arc length (percentage shown) of each segment is proportional to the chromosome's relative fitness, computed as the inverse of its total operation time (evaluation value, in minutes): chromosomes with lower operation times receive larger arc lengths and thus higher selection probabilities. The weighted, non-uniform nature of the wheel, as opposed to a uniform wheel where all segments are equal, ensures that fitter individuals are more likely to be selected as parents for crossover, while still preserving diversity by allowing lower-fitness individuals a non-zero selection chance.}
    \label{roulette}
\end{figure}

\subsubsection{Selection}
To guide the evolutionary search toward fitter solutions, we employ the \textit{Roulette Wheel Selection} method, a probabilistic selection strategy based on fitness-proportional sampling. In this approach, each chromosome in the population is assigned a selection probability proportional to its fitness (or inversely proportional to its cost, depending on the objective). The method draws inspiration from the spinning mechanism of a roulette wheel, where segments correspond to individuals and the size of each segment reflects their relative fitness. Unlike uniform roulette models, where all candidates have equal chances, our implementation uses a \textit{weighted} roulette wheel (see Figure~\ref{roulette}), where fitter individuals occupy larger segments, thereby increasing their likelihood of being selected for reproduction.

\paragraph{Elite Preservation}
To maintain strong genetic material across generations, we apply \textit{elitism}, a strategy that guarantees survival of the top-performing individuals. Let $P_E$ denote the elite percentage of the population per generation. These elite chromosomes are automatically copied to the next generation without undergoing crossover or mutation. In our implementation, we set $P_E = 20\%$.

\paragraph{Notations}
$P_E$ denotes the percentage of elite class chromosomes of a generation. $E_{rw}$ denotes the end value of the roulette wheel.  

The detailed steps of the roulette wheel selection procedure are provided in Algorithm~\ref{alg:roulette}.

\begin{algorithm}[!ht]
    \caption{Roulette Wheel Selection Algorithm}
    \label{alg:roulette}
    \KwIn{Probability against the fitness value of each chromosome}
    \KwOut{A selected chromosome}

    Define a 1D vector $RW$ of size $n$ for storing the fitness value of each chromosome. The fitness value is stored as a \textit{cumulative sum} order where $E_{rw}$ is the total sum of all fitness.

    \For{$i \leftarrow 1$ \KwTo $n$}{
        Generate a random number $r$, where $0\leq r\leq E_{rw}$.

        Select a chromosome as a parent for crossover.
    }
\end{algorithm}

\subsubsection{Termination}
The termination condition of the GA determines when the run ends. Initially, the GA progresses quickly, yielding better solutions every few iterations. However, this progress tends to slow down later, with minimal improvements. To guarantee that our solution approaches optimality, we establish a termination condition as follows: $g_i$ denotes the $i^{th}$ generation, $G$ denotes the maximum number of generations, and $N_s$ denotes the number of successive generations in which the fittest chromosome incurs the same cost. The GA execution concludes according to the criteria specified in Algorithm \ref{alg:termination}.

\begin{algorithm}[H]
    \caption{GA termination algorithm}
    \label{alg:termination}
    \KwIn{Number of successive generations in which the cost of the fittest chromosome is the same and the iteration number}
    \KwOut{Boolean value to take termination decision}
    
    $N_s \leftarrow$ Number of successive generations in which the fittest chromosome costs the same.
    
    \If{$g_i=G$ \textbf{or} $N_s = 100$} {
        Terminate the GA run.
    }
    \Else{
        Continue
    }
\end{algorithm}

\subsubsection{Parameters}
The GA control parameters are shown in Table \ref{tab:GA_parameters}. The parameters best fit our algorithm, such as population size, crossover technique, elite percentage, mutation probability, selection method, etc. As the solution to our problem is a smooth landscape type and the complexity of our problem is medium, we selected these parameters to fit the situation.

\begin{table}[!ht]
\centering
\caption{GA control parameters}
\label{tab:GA_parameters}
\begin{tabular}{l l l} 
    \toprule[1.5pt]
    \textbf{Parameter} & \textbf{Search Space} & \textbf{Selected Value}\\ 
    \midrule[1pt]
    Population size & [50, 300]                   & 200\\
    1D crossover strategy & One-Point, Uniform, Two-Point & Two-Point Crossover\\
    2D crossover strategy & Random Block, 2D Substring & 2D Substring\\ 
    Crossover rate        & [0.6, 0.9]             & 0.8\\
    1D mutation strategy & Inversion, Swap & Swap\\
    2D mutation strategy & Random-cell, 2D Two-point Swap & 2D Two-point Swap\\
    Mutation rate         & [0.1, 0.5]            & 0.3\\
    Selection strategy    & Tournament, SUS, Roulette             & Roulette\\
    Elite class & [0.05, 0.3] & 0.2\\
    Consecutive iterations & [500, 2000] & 1000\\
    \bottomrule[1.5pt]
    \end{tabular}
\end{table}

We began by evaluating several population sizes (e.g., 50, 100, 200, and 300). A size of 200 was chosen because it consistently yielded better-quality solutions at a manageable computational cost. Smaller populations converged faster but often prematurely, while larger sizes increased runtime without proportional improvements. For the 1D chromosome, we selected the Two-Point Crossover, as it preserves contiguous gene blocks and encourages structural inheritance, proving more effective than single-point and uniform crossover in our tests. For the 2D chromosome, we used a 2D Substring Crossover, which accounts for the spatial relationships among genes, which are crucial in applications such as layout optimization or scheduling. The crossover rate was tested in the range of 0.6 to 0.9. We selected 0.80 based on its ability to maintain a strong diversity-exploitation balance, outperforming more conservative rates that hindered exploration. We applied a Swap Mutation to 1D chromosomes and a custom 2D Two-Point Swapping Mutation to 2D chromosomes. The mutation rate of 0.30 was selected after trials with 0.1 to 0.5; higher rates injected necessary variability without destabilizing the population. For selection, we experimented with Tournament Selection and Stochastic Universal Sampling but ultimately adopted Roulette Wheel Selection, which provides proportional selection pressure and performed more consistently for our problem type. An elite class size of 20\% was used to retain top-performing individuals between generations. Tests with 5\%, 10\%, and 30\% revealed that 20\% offered the best convergence reliability without hindering innovation. Finally, we ran preliminary experiments with various stopping criteria. We found that 1000 consecutive iterations offered a practical balance between convergence likelihood and runtime.

\section{Results}
\label{results}
This section addresses the magnitude of \textit{QCDC-DR-GA}. We provide tools to translate cycle-based benefits into time equivalents and validate those estimates against real-world double-cycling data. With an eye on the present and future, we analyze the financial impact of double cycling, estimating the potential rewards for both existing ships and those that will grace the waves in the years ahead. The experiments were run on a computer with 8GB of RAM, Ubuntu 22.04, and an Intel Core i5 8th Gen. The algorithm was implemented using Python libraries- Pandas and NumPy.

\subsection{Performance Comparisons of the Algorithms}
\label{com_algorithms}
To validate the effectiveness of our proposed QCDC-DR-GA algorithm, we conducted a comprehensive comparison with three established optimization methods:
\begin{enumerate}[label=(\alph*)]
    \item \textit{Dual-Cycling Greedy Upper Bound Approach:} It represents a greedy approach in which the unloading sequence is generated by sorting the stacks of containers of the ship row in descending order \citep{goodchild2006double}. This heuristic method focuses solely on implementing dual-cycle loading/unloading, neglecting dockside rehandles. It's a baseline approach to gauge the potential improvement offered by more complex methods.
    \item \textit{Mixed-Integer Programming Algorithm for QCDCS (bi-level GA):} This approach represents an improvement over the greedy upper bound by incorporating QCDCS optimization within a bi-level GA framework \citep{zeng2015simulation}. It came with a significant advancement over the greedy approach by incorporating QCDCS optimization.
    \item \textit{GA-ILSRS:} This method optimized dockyard rehandles using a GA combined with Iterated Local Search \citep{ji2015optimization}. This work only focused on dockyard rehandles, and ours is a combination of optimizing dockyard rehandles and dual cycling of container loading-unloading. Hence, for container unloading and loading, the following two scenarios are possible.
    \begin{enumerate}[label=(\roman*)]
        \item \textit{Scenario 1:} Considers dual cycling for loading/unloading while neglecting dockyard rehandles.
        \item \textit{Scenario 2:} Focuses solely on dockyard rehandling with single-cycle loading/unloading.
    \end{enumerate}
\end{enumerate}

The greedy upper bound is a heuristic that implements dual cycling only, but it does not consider any dockside rehandling. The QCDCS-bilevel GA overcame this heuristic approach but did not account for the time required for dockyard rehandles. On the other hand, the GA-ILSRS only optimized the dockyard rehandles without considering the loading-unloading system. As for QCDC-DR-GA, we combined loading-unloading with dual cycling and dockyard rehandles into a single algorithm and optimized it using a sophisticated GA approach.

\subsection{Datasets}\label{dataset}
Six scenarios were designed based on the number of stacks and the maximum container height in each row. Given the characteristics of commonly used container ships, we assume the number of stacks is 5-30, and the maximum stack height is 4-10. This limit is considered based on real-time ship sizes (Table \ref{tab:dataset}). Here, the large size ships can carry 18,000 to 22,000 containers, the medium size ships can carry 11,000 to 17,999 containers, and the small size ships can carry 4,500 to 10,999 containers \citep{ship_size}. The program generated a loading and unloading plan for one row and the dockyard container arrangements. Table \ref{tab:dataset} shows the configurations of six datasets of unloading and loading plans. Table \ref{table3} shows a sample unloading plan of a small ship, and Table \ref{table2} shows a sample loading plan of that ship.

\begin{table*}[!ht]
\centering
\caption{Loading-unloading plan configuration}
\label{tab:dataset}
\begin{tabular}{cccc} 
\toprule[1.5pt]
\textbf{Scenario} & \textbf{Ship Size} & \textbf{No. of Stacks in a Row} & \textbf{Maximum Stack Height}  \\ 
\midrule[1pt]
\textbf{1}   &  Large   & 30                     & 10
                \\
\textbf{2}   & Large    & 25                     & 10                             \\
\textbf{3}   & Medium     & 20                     & 10                             \\
\textbf{4}   & Medium     & 15                     & 8                              \\
\textbf{5}   & Small     & 10                     & 5                              \\
\textbf{6}    & Small    & 5                      & 4                              \\
\bottomrule[1.5pt]
\end{tabular}
\end{table*}

Each cell tells the container location information in Table \ref{table3}. For example, $1B$ means \rh{the container destined for the 1st stack and 2nd tier of the row (tier alias B~$=$~Tier~2), following the global \texttt{[stack][tier]} convention,} and $F$ means the container would remain on the ship. The information in Table \ref{table2} is similar. 

\begin{table*}[!ht]
\centering
\caption{\rh{Unloading plan of a ship's row, displayed with \textbf{Tier No.\ as rows} and \textbf{Stack No.\ as columns} to match the physical stacking orientation (tiers increase upward, Tier~1 bottom-most). Each non-empty cell holds a container identifier in the \texttt{[stack][tier]} convention of Section~\ref{problem}; for example, `\texttt{2B}' is the container destined for ship stack~2 at tier~2. Cells labeled `F' denote containers that remain on the ship; empty cells indicate no container at that position.}}
\label{table3}
\begin{tabular}{c|cccccccccc}
\toprule[1.5pt]
\diagbox{\textbf{Tier No.}}{\textbf{Stack No.}} 
  & \textbf{1} & \textbf{2} & \textbf{3} & \textbf{4} & \textbf{5}
  & \textbf{6} & \textbf{7} & \textbf{8} & \textbf{9} & \textbf{10} \\
\midrule[1pt]
\textbf{1 (bottom)} & 1A & 2A & F  & 4A & 5A & F  & F  & F  & F  & F   \\
\textbf{2}          & 1B & 2B & 3A & 4B & 5B & 6A & 7A & 8A & 9A & 10A \\
\textbf{3}          & 1C & 2C & 3B &    & 5C & 6B & 7B & 8B & 9B & 10B \\
\textbf{4}          & 1D &    & 3C &    & 5D & 6C & 7C & 8C & 9C & 10C \\
\textbf{5 (top)}    & 1E &    & 3D &    & 5E & 6D & 7D & 8D & 9D & 10D \\
\bottomrule[1.5pt]
\end{tabular}
\end{table*}

\begin{table*}[!ht]
\centering
\caption{\rh{Loading plan of a ship's row, displayed with \textbf{Tier No.\ as rows} and \textbf{Stack No.\ as columns}, consistent with Table~\ref{table3} (Tier~1 bottom-most, containers loaded upward). Each cell holds a container identifier in the same \texttt{[stack][tier]} convention; empty cells indicate tier positions that remain unoccupied after loading.}}
\label{table2}
\begin{tabular}{c|cccccccccc}
\toprule[1.5pt]
\diagbox{\textbf{Tier No.}}{\textbf{Stack No.}}
  & \textbf{1} & \textbf{2} & \textbf{3} & \textbf{4} & \textbf{5}
  & \textbf{6} & \textbf{7} & \textbf{8} & \textbf{9} & \textbf{10} \\
\midrule[1pt]
\textbf{1 (bottom)} & 1A  & 2A  & 3A  & 4A  & 5A  & 6A  & 7A  & 8A  & 9A  & 10A \\
\textbf{2}          & 1B  & 2B  & 3B  & 4B  & 5B  & 6B  & 7B  &     & 9B  & 10B \\
\textbf{3}          & 1C  & 2C  &     & 4C  & 5C  &     & 7C  &     & 9C  & 10C \\
\textbf{4}          & 1D  & 2D  &     & 4D  & 5D  &     & 7D  &     & 9D  & 10D \\
\textbf{5 (top)}    & 1E  & 2E  &     & 4E  & 5E  &     &     &     &     &     \\
\bottomrule[1.5pt]
\end{tabular}
\end{table*}

The program also generated the dockyard plan according to the loading plan. Here, we assumed that the highest dockyard container stack height is 6. The sample of the dockyard plan is elaborated in \ref{sec:initial_pop} as $A_2$.
    
\begin{table*}[hbt!]
\centering
\caption{Two-tailed paired t-test results of the proposed QCDC-DR-GA against the other strategies}
\label{tab:t-test}
\resizebox{\textwidth}{!}{%
\begin{tabular}{c c c c ccccccc c c}
\toprule[1.5pt]
\multirow{2}{*}{\thead{\textbf{Scenarios}}}&
  \multirow{2}{*}{\textbf{\thead{No.\\of\\Stacks}}} &
  \multirow{2}{*}{\textbf{\thead{Maximum\\Stack\\Height}}} &
  \multirow{2}{*}{\thead{\textbf{Strategies}}} &
  \multicolumn{7}{c}{\thead{\textbf{Operation Time (min)}}} &
  \textbf{$t_{19}(0.05)$ = 2.093} &
  \multirow{2}{*}{\textbf{\thead{Improvement of\\QCDC-DR-GA (\%)}}} \\ \cline{5-11}
 &
   &
   &
   &
  \thead{\textbf{Min}} &
  \thead{\textbf{Max}} &
  \thead{\textbf{Mean}} &
  \textbf{\thead{Standard\\deviation}} &
  \textbf{\thead{Pearson\\correlation\\coefficient\\$(r)$}} &
  \thead{\textbf{t-statistic}} &
  \thead{\textbf{p-value}} &
  \thead{\textbf{Significance}} &
   \\ \midrule[1pt]
   
\multirow{5}{*}{1} & \multirow{5}{*}{30} & \multirow{5}{*}{10} & Proposed QCDC-DR-GA & 615.25 & 704.92 & 667.62 & 25.94 & -         & -           & -         & -   & -       \\
                   &                     &                     & Greedy Upper Bound  & 718.58 & 825.92 & 780.33 & 35.95 & -0.16317  & -10.580512  & 2.10E-09  & Yes & 14.44\% \\
                   &                     &                     & QCDC-bi-level-GA    & 714.92 & 813.25 & 763.00 & 32.07 & -0.24243  & -9.298231   & 1.68E-08  & Yes & 12.50\% \\
                   &                     &                     & GA-ILSRS-Scenario-1 & 669.25 & 764.25 & 708.95 & 28.92 &  0.07729  & -4.952353   & 8.84E-05  & Yes & 5.83\%  \\
                   &                     &                     & GA-ILSRS-Scenario-2 & 820.75 & 851.25 & 835.09 &  9.84 & -0.25862  & -24.938504  & 5.57E-16  & Yes & 20.05\% \\
\midrule[0.8pt]
\multirow{5}{*}{2} & \multirow{5}{*}{25} & \multirow{5}{*}{10} & Proposed QCDC-DR-GA & 522.42 & 604.75 & 559.02 & 27.95 & -         & -           & -         & -   & -       \\
                   &                     &                     & Greedy Upper Bound  & 645.33 & 710.75 & 674.20 & 21.47 & -0.00945  & -14.548904  & 9.41E-12  & Yes & 17.08\% \\
                   &                     &                     & QCDC-bi-level-GA    & 652.17 & 722.75 & 672.10 & 23.10 &  0.45033  & -18.675141  & 1.10E-13  & Yes & 16.82\% \\
                   &                     &                     & GA-ILSRS-Scenario-1 & 535.42 & 631.08 & 579.37 & 29.01 &  0.07522  & -2.349253   & 2.98E-02  & Yes & 3.51\%  \\
                   &                     &                     & GA-ILSRS-Scenario-2 & 666.00 & 691.00 & 680.79 &  8.32 & -0.26691  & -17.447414  & 3.75E-13  & Yes & 17.89\% \\
\midrule[0.8pt]
\multirow{5}{*}{3} & \multirow{5}{*}{20} & \multirow{5}{*}{10} & Proposed QCDC-DR-GA & 381.42 & 452.08 & 417.35 & 23.09 & -         & -           & -         & -   & -       \\
                   &                     &                     & Greedy Upper Bound  & 491.08 & 563.08 & 523.60 & 20.17 & -0.37587  & -13.229620  & 4.90E-11  & Yes & 20.29\% \\
                   &                     &                     & QCDC-bi-level-GA    & 484.08 & 563.08 & 519.85 & 26.43 & -0.33001  & -11.338757  & 6.71E-10  & Yes & 19.72\% \\
                   &                     &                     & GA-ILSRS-Scenario-1 & 397.42 & 465.75 & 425.77 & 20.47 &  0.25918  & -1.415272   & 1.73E-01  & No  & 1.98\%  \\
                   &                     &                     & GA-ILSRS-Scenario-2 & 492.25 & 511.50 & 502.76 &  5.10 & -0.13529  & -15.710274  & 2.43E-12  & Yes & 16.99\% \\
\midrule[0.8pt]
\multirow{5}{*}{4} & \multirow{5}{*}{15} & \multirow{5}{*}{8}  & Proposed QCDC-DR-GA & 201.67 & 259.00 & 229.01 & 19.34 & -         & -           & -         & -   & -       \\
                   &                     &                     & Greedy Upper Bound  & 295.33 & 358.67 & 327.83 & 19.39 & -0.29172  & -14.199917  & 1.44E-11  & Yes & 30.14\% \\
                   &                     &                     & QCDC-bi-level-GA    & 295.50 & 349.33 & 323.00 & 18.00 &  0.01337  & -16.016549  & 1.73E-12  & Yes & 29.10\% \\
                   &                     &                     & GA-ILSRS-Scenario-1 & 211.33 & 277.67 & 248.65 & 18.25 &  0.51096  & -4.719124   & 1.49E-04  & Yes & 7.90\%  \\
                   &                     &                     & GA-ILSRS-Scenario-2 & 279.00 & 295.00 & 286.56 &  5.66 & -0.12894  & -12.349324  & 1.59E-10  & Yes & 20.08\% \\
\midrule[0.8pt]
\multirow{5}{*}{5} & \multirow{5}{*}{10} & \multirow{5}{*}{5}  & Proposed QCDC-DR-GA &  89.50 & 138.67 & 119.40 & 16.81 & -         & -           & -         & -   & -       \\
                   &                     &                     & Greedy Upper Bound  & 138.50 & 186.67 & 164.44 & 15.05 & -0.09791  & -8.523389   & 6.46E-08  & Yes & 27.39\% \\
                   &                     &                     & QCDC-bi-level-GA    & 147.00 & 183.67 & 162.13 & 12.82 & -0.51102  & -7.397294   & 5.26E-07  & Yes & 26.36\% \\
                   &                     &                     & GA-ILSRS-Scenario-1 &  97.00 & 147.67 & 123.13 & 15.96 & -0.06245  & -0.698810   & 4.93E-01  & No  & 3.03\%  \\
                   &                     &                     & GA-ILSRS-Scenario-2 & 137.00 & 152.00 & 145.76 &  4.64 &  0.29488  & -7.337783   & 5.90E-07  & Yes & 18.09\% \\
\midrule[0.8pt]
\multirow{5}{*}{6} & \multirow{5}{*}{5}  & \multirow{5}{*}{4}  & Proposed QCDC-DR-GA &  37.67 &  70.42 &  55.43 & 11.03 & -         & -           & -         & -   & -       \\
                   &                     &                     & Greedy Upper Bound  &  48.67 &  81.42 &  63.82 & 13.20 &  0.06150  & -2.250737   & 3.64E-02  & Yes & 13.15\% \\
                   &                     &                     & QCDC-bi-level-GA    &  47.67 &  80.42 &  65.60 & 11.03 &  0.06415  & -3.015466   & 7.11E-03  & Yes & 15.51\% \\
                   &                     &                     & GA-ILSRS-Scenario-1 &  37.67 &  70.42 &  53.84 & 10.84 & -0.20072  &  0.417813   & 6.81E-01  & No  & $-$2.94\% \\
                   &                     &                     & GA-ILSRS-Scenario-2 &  55.50 &  66.00 &  60.05 &  3.13 &  0.35232  & -1.998582   & 6.02E-02  & No  & 7.70\%  \\ \bottomrule[1.5pt]
\end{tabular}%
}
\end{table*}

\subsection{Numerical Tests}
The six scenarios described in subsection \ref{dataset}, Table \ref{tab:dataset} are used for numerical tests. The processing time of QC for single and dual cycling is taken from the trial run done by \cite{goodchild2006double}, mentioned in subsection \ref{dualcycle}, is \rh{105} seconds and 170 seconds. The rehandling time of a container for a gantry crane at the dockyard is generated from a uniform distribution of 60 seconds. 

\subsubsection{Test Results}
\begin{figure*}[!ht]
\begin{subfigure}[b]{0.45\linewidth}
    \centering
    \includegraphics[width=\linewidth]{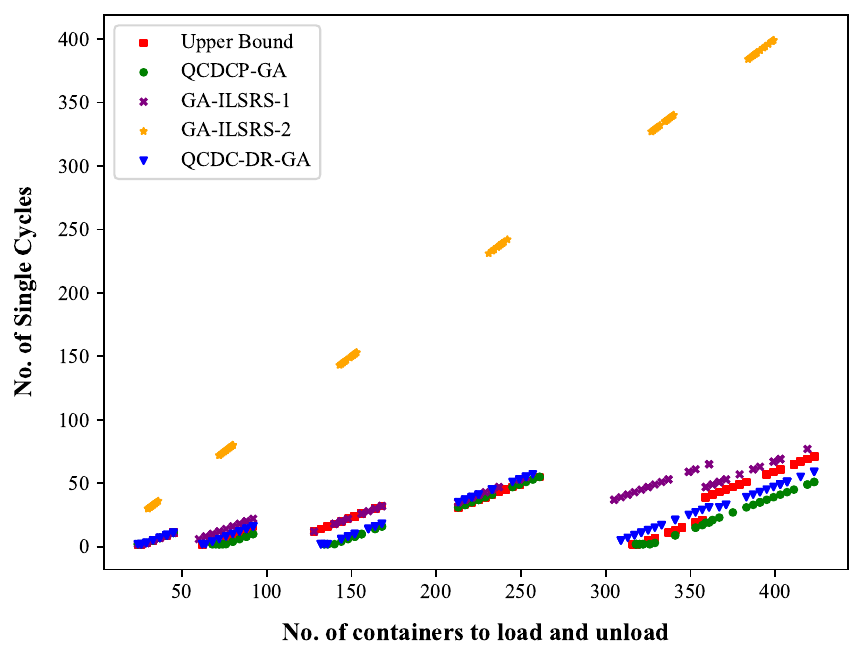}
    \caption{Performance comparison on minimizing the number of SCs of QCDC-DR-GA and the other four algorithms}
    \label{fig:result_graph_SC}
\end{subfigure}
\hfill
\begin{subfigure}[b]{0.45\linewidth}
    \centering
    \includegraphics[width=\linewidth]{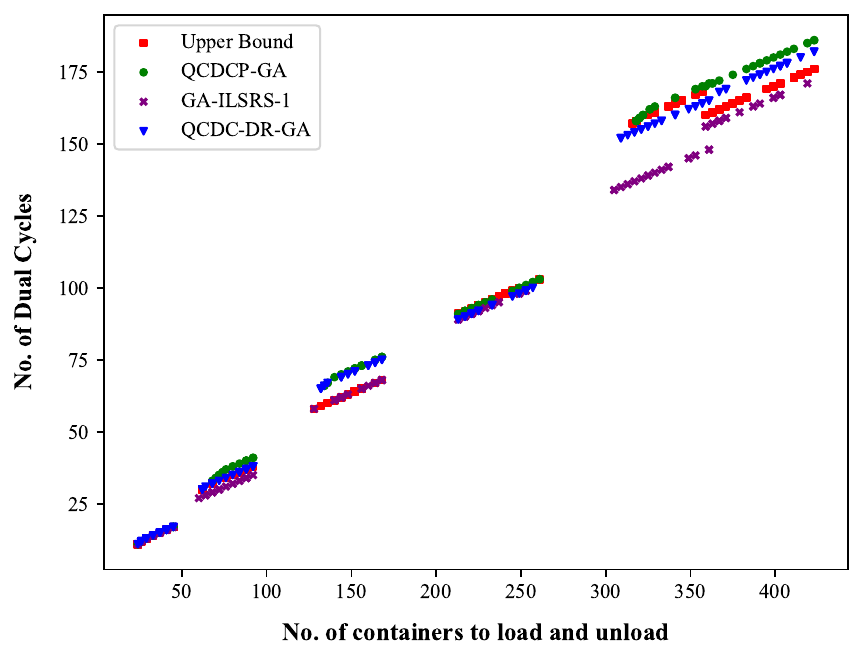}
    \caption{Performance comparison on maximizing the number of DCs of QCDC-DR-GA and the other four algorithms}
    \label{fig:result_graph_DC}
\end{subfigure}
\hfill
\begin{subfigure}[b]{0.45\linewidth}
    \centering
    \includegraphics[width=\linewidth]{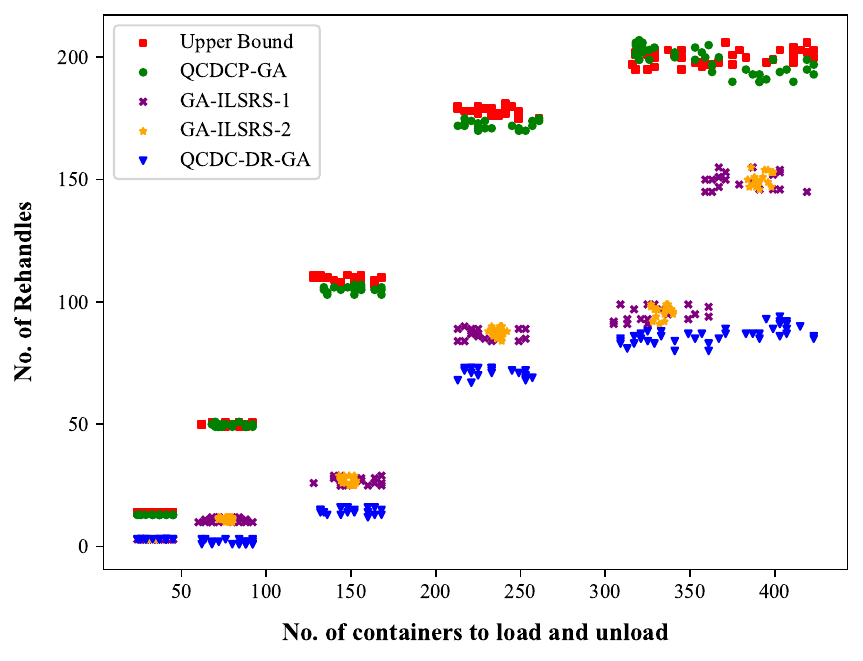}
    \caption{Performance comparison on minimizing the number of rehandles of QCDC-DR-GA and the other four algorithms}
    \label{fig:result_graph_rehandles}
\end{subfigure}
\hfill
\begin{subfigure}[b]{0.45\linewidth}
    \centering
    \includegraphics[width=\linewidth]{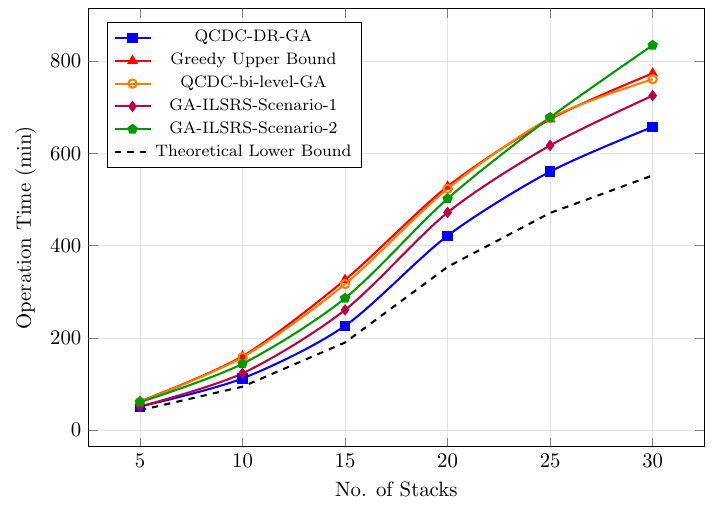}
    \caption{Final Performance comparison of QCDC-DR-GA and other four algorithms}
    \label{result_graph}
\end{subfigure}
\caption{\rh{Comprehensive performance comparison of QCDC-DR-GA with four other algorithms across four key metrics: (a)~minimization of single cycles (SCs), (b)~maximization of dual cycles (DCs), (c)~minimization of dockyard rehandles, and (d)~total operation time, the primary optimization objective. The baseline labeled \emph{Greedy Upper Bound} in every panel is the QCDCS Greedy heuristic of} \citet{goodchild2006double}\rh{, which greedily maximizes dual cycles without considering rehandles; it appears as a reference curve in all four panels. Panel~(d) additionally plots the theoretical lower bound defined in the main text. Operation time is reported in minutes.}}
\end{figure*}

We comprehensively evaluated the proposed QCDC-DR-GA algorithm by comparing its performance with four baseline methods on six datasets described in Table \ref{tab:dataset}, each representing distinct scenarios with varying container counts and ship configurations. The detailed simulation outcomes are presented in Table \ref{tab:t-test}, while the graphical illustrations of performance are provided in Figure \ref{fig:result_graph_SC}, \ref{fig:result_graph_DC}, \ref{fig:result_graph_rehandles}, and \ref{result_graph}.

{As observed in Figure~\ref{fig:result_graph_SC} and Figure~\ref{fig:result_graph_DC}, certain algorithms, especially QCDC-bi-level-GA and GA-ILSRS-Scenario-1, achieve a slightly better count on individual metrics (number of SCs minimized or DCs maximized, respectively) in some scenarios, as those methods are explicitly designed to optimize that single metric. Likewise, Figure~\ref{fig:result_graph_rehandles} shows that algorithms focused solely on rehandle minimization (GA-ILSRS-Scenario-1, GA-ILSRS-Scenario-2) do not consistently outperform QCDC-DR-GA even on that individual metric, because QCDC-DR-GA's joint optimization implicitly constrains rehandles while also controlling cycle efficiency.

The key objective of this study is to holistically minimize the total operation time of the container loading and unloading process, which combines all three components: single cycles, dual cycles, and rehandles. Figure~\ref{result_graph} delivers the most comprehensive evaluation. QCDC-DR-GA achieves the lowest mean total operation time in all six scenarios, with improvements over the best-performing single-objective baseline of 5.8--20.1\% for large ships (Scenarios~1--2) and 2.0--30.1\% across all scenarios. These differences are statistically significant (paired t-test, $\alpha = 0.05$) in 20 of 24 pairwise comparisons. The four non-significant comparisons occur in the smallest ship configurations (Scenarios~3, 5, and~6), where absolute time differences are small, and QCDC-DR-GA is not worse than any baseline by a statistically significant margin. The performance advantage of QCDC-DR-GA grows monotonically with ship size (number of stacks per row), consistent with the hypothesis that the benefit of joint optimization scales with problem complexity.}

\rh{To place these results on an absolute scale, Figure~\ref{result_graph} also plots a \emph{theoretical lower bound}, defined as the operation time of an idealized schedule in which every crane move is a dual cycle (no single cycles, $w_s=0$) and the dockyard requires no rehandles ($R=0$). Under the cost model $T=\alpha w_s+\beta w_d+\gamma R$ this reduces to $T_{\mathrm{lb}}=\beta\,w_d^{\min}$, where $w_d^{\min}$ is the smallest number of dual cycles required to discharge and load the given plan, equivalently the number of paired unload--load moves dictated by the loading manifest. Reading the achieved curves against this bound shows that, for the large-ship scenarios (Scenarios~1--2), QCDC-DR-GA operates approximately 15--20\% above the theoretical lower bound, whereas the Greedy Upper Bound and the single-objective GA baselines remain roughly 30--45\% above it. The gap between QCDC-DR-GA and the lower bound narrows as ship size grows, indicating that the joint optimization recovers an increasing share of the theoretically available time savings precisely in the large, complex configurations where those savings matter most. The residual gap is attributable to structural features of the loading that manifest as a non-zero number of single cycles and unavoidable rehandles, so the lower bound itself is not achievable in practice; it serves only as a fixed reference for interpreting the achieved times. All operation-time values in this discussion and in Figure~\ref{result_graph} are reported in minutes, consistent with Table~\ref{tab:t-test}.}

{
\subsubsection{Key Findings}
The key findings and remarks of the simulation are as follows:
\begin{enumerate}
    \item QCDC-DR-GA achieves statistically significant improvements over all four baselines in 20 out of 24 pairwise comparisons (paired t-test, $\alpha = 0.05$). The four non-significant cases all involve GA-ILSRS-Scenario-1 or GA-ILSRS-Scenario-2 in small-ship scenarios (Scenarios 3, 5, and 6), where the overall operation time is short enough that the absolute differences fall below the threshold of statistical detectability with $n = 20$ instances. Importantly, the point estimate of QCDC-DR-GA's mean operation time remains lower than or equal to the comparator in all but one case (Scenario~6 vs.\ GA-ILSRS-Scenario-1, where the 2.9\% difference is also not significant), indicating no evidence that QCDC-DR-GA is worse than any baseline.
    \item The results support the hypothesis that separately optimizing QCDC and dockyard rehandles leads to suboptimal outcomes. Compared to QCDC-DR-GA, methods such as QCDC Scheduling Optimized by bi-level GA and GA-ILSRS (Scenario 2) show only limited improvement due to their fragmented approach; for large ships (Scenarios 1--2), QCDC-DR-GA outperforms the best single-objective baseline by 12.5--20.1\%.
    \item Similarly, neglecting dual-cycling in QC operation optimization, as in GA-ILSRS (Scenario 1), leads to inferior performance compared to QCDC-DR-GA in all scenarios where the difference is statistically significant. Critically, the joint optimization of A2 within QCDC-DR-GA reduces dockyard rehandles by 11--84\% relative to the DC-optimized-only baseline (GA-ILSRS-Scenario-1), confirming that the co-optimization of the dockyard layout is a primary source of performance gain in larger ship configurations.
    \item The improvement of QCDC-DR-GA over all baselines grows consistently with ship size (number of stacks per row). For Scenario~1 (30 stacks, large ship), improvements over the best single-objective method reach 20.1\%, while for Scenario~6 (5 stacks, small ship) the advantage narrows to 7.7--15.5\% (excluding the one non-significant comparison). This scaling behavior is consistent with the hypothesis that integrated optimization becomes increasingly beneficial as the combinatorial complexity of the joint problem grows.
\end{enumerate}
}
By integrating QCDC optimization and dockyard rehandle minimization into a single algorithm, QCDC-DR-GA demonstrates significant performance advantages over existing methods. This unified approach offers a powerful tool for port operators seeking to minimize container handling times and maximize operational efficiency.

{
\subsection{Ablation: Fixed vs. Optimized Initial Configuration}
\label{sec:ablation}

A key concern in evaluating QCDC-DR-GA is whether its performance advantage stems primarily from its inherent algorithmic search capability (its optimization of the unloading sequence $A_1$) or from the additional benefit of co-optimizing the initial dockyard configuration $A_2$ alongside $A_1$. \rh{To disentangle these two sources of improvement rigorously, we report two complementary analyses: (i)~a \emph{controlled fixed-$A_2$ experiment}, in which the \emph{same} QCDC-DR-GA algorithm is run with $A_2$ frozen throughout the GA loop, so that the only difference between the two arms is whether $A_2$ is optimized; and (ii)~a \emph{proxy cross-check} using the existing baseline data. Analysis~(i) is the decisive isolation of the $A_2$ contribution because it varies a single factor within one algorithm; analysis~(ii) is retained only as an independent, corroborating comparison.}

\rh{%
\subsubsection{Controlled Fixed-$A_2$ Experiment}
\label{sec:controlled_a2}
We define a controlled ablation variant, denoted \textbf{QCDC-DR-GA(fixed-$A_2$)}, that is byte-for-byte identical to the full QCDC-DR-GA in every respect, the same 1D chromosome ($A_1$) encoding, the same two-point 1D crossover and swap mutation, the same roulette selection, the same elitism, the same population size and stopping rule, and the same fitness function $T=\alpha w_s+\beta w_d+\gamma R$, with a single modification: the dockyard plan $A_2$ is generated once at initialization (using the random nearest-lowest heuristic) and then \emph{frozen} for the entire run. Concretely, the 2D crossover and 2D mutation operators that act on $A_2$ are disabled, so $A_2$ is never altered during evolution, while $A_1$ continues to evolve exactly as in the full method. Comparing QCDC-DR-GA against QCDC-DR-GA(fixed-$A_2$) therefore isolates the contribution of $A_2$ co-optimization while holding the scheduling search and all hyper-parameters constant; the difference between the two arms cannot be attributed to any factor other than whether $A_2$ is optimized.

Both arms were executed under the identical experimental protocol of Section~\ref{dataset} ($n=20$ independent runs per scenario; cost weights $\alpha=105$\,s, $\beta=170$\,s, $\gamma=60$\,s). \rh{Because the instance generator used for Table~\ref{tab:t-test} was not seeded and its problem instances were not archived, they cannot be recovered exactly; we therefore generated a single fixed instance set from a fixed random seed, archived it, and released it with this paper, and ran both arms of the ablation paired on that set so that the experiment is fully reproducible. As a consistency check, the Opt-$A_2$ column below reproduces the QCDC-DR-GA means of Table~\ref{tab:t-test} to within about $2\%$ on the large-ship scenarios, where the run-to-run variance is small; the wider difference on Scenarios~5 and~6 reflects the high relative variance of tiny-ship instances (Table~\ref{tab:t-test} reports standard deviations of $16.81$ and $11.03$ minutes for these scenarios, i.e.\ coefficients of variation of $14$--$20\%$), so an independent draw of $20$ ships shifts the mean by a comparable amount. The difference is thus a sampling consequence of the original instances not having been archived, not a difference in method.} Table~\ref{tab:fixed_a2} reports the results for the small-ship scenarios (Scenarios~5 and~6), for which the controlled experiment is most directly informative about robustness under limited yard reconfiguration, the operating regime in which fixing $A_2$ is the binding constraint.

\begin{table}[!ht]
\centering
\caption{\rh{Controlled fixed-$A_2$ ablation: the \emph{same} QCDC-DR-GA run with $A_2$ co-optimized versus $A_2$ frozen throughout the GA loop, paired on one archived, seeded instance set ($n=20$ runs per scenario). Operation time in minutes; rehandles are the mean dockyard rehandle count. Both arms are costed with the same weights as Table~\ref{tab:t-test} ($\alpha=105$\,s, $\beta=170$\,s, $\gamma=60$\,s). Significance by two-tailed paired $t$-test ($\alpha=0.05$).}}
\label{tab:fixed_a2}
\begin{tabular}{c c c c c c c}
\toprule[1.5pt]
\textbf{Scenario} & \textbf{Stacks} & \makecell{\textbf{Opt-$A_2$}\\\textbf{time}} & \makecell{\textbf{Fixed-$A_2$}\\\textbf{time}} & \makecell{\textbf{Time}\\\textbf{impr.}} & \makecell{\textbf{Rehandle}\\\textbf{reduction}} & \textbf{Sig.} \\
\midrule[1pt]
\rh{5} & \rh{10} & \rh{111.63} & \rh{116.20} & \rh{3.9\%} & \rh{69.6\%} & \rh{Yes ($p<10^{-7}$)} \\
\rh{6} & \rh{5}  & \rh{41.38}  & \rh{44.45}  & \rh{6.9\%} & \rh{98.1\%} & \rh{Yes ($p<10^{-7}$)} \\
\bottomrule[1.5pt]
\end{tabular}
\end{table}

Two conclusions follow directly. First, even in these small configurations, the full QCDC-DR-GA significantly outperforms its own fixed-$A_2$ variant in total operation time, and it removes essentially all of the dockyard rehandles that the fixed-$A_2$ variant incurs; the gains are statistically significant under the paired $t$-test. \rh{Second, and crucially for practical robustness, the fixed-$A_2$ arm remains close to the optimized arm: freezing $A_2$ costs only $3.9\%$ and $6.9\%$ of operation time in Scenarios~5 and~6 respectively, so the bulk of QCDC-DR-GA's performance is retained even when the yard cannot be reconfigured. Because Table~\ref{tab:fixed_a2} and Table~\ref{tab:t-test} are separate experiments (run on independently drawn instance sets), we deliberately restrict this conclusion to the paired within-table comparison and do not compare the fixed-$A_2$ times against the external baselines of Table~\ref{tab:t-test}.} The advantage of $A_2$ co-optimization is therefore an \emph{added} benefit on top of an already-competitive scheduler, not the sole source of QCDC-DR-GA's performance.
}

\subsubsection{Proxy Cross-Check Using Existing Experimental Data}
\rh{As an independent corroboration of the controlled experiment above,} we additionally exploit the structure of the existing baselines. Specifically, GA-ILSRS-Scenario-1 optimizes the unloading sequence $A_1$ for dual-cycle performance using a metaheuristic, but its dockyard arrangement is generated by a random nearest-lowest-stack heuristic and is never updated during the optimization run; i.e., $A_2$ is effectively fixed and unoptimized. This makes GA-ILSRS-Scenario-1 a useful external approximation to a fixed-$A_2$ condition. \rh{We stress that, unlike the controlled experiment of Section~\ref{sec:controlled_a2}, this comparison involves two different algorithms and therefore conflates the scheduling mechanism with the $A_2$ treatment; it is reported only as a complementary check, and the controlled experiment is the authoritative isolation of the $A_2$ effect.}

\subsubsection{Results and Interpretation}
\rh{The two analyses agree.} \rh{For the larger ships, where the controlled ablation was not run, we rely on the proxy cross-check of the previous subsection (QCDC-DR-GA versus the fixed-$A_2$ external baseline GA-ILSRS-Scenario-1); for the small ships we use the direct controlled measurement of Table~\ref{tab:fixed_a2}.} Comparing QCDC-DR-GA (joint $A_1$+$A_2$) against the fixed-$A_2$ condition, the benefit of $A_2$ co-optimization scales strongly with ship size:

\begin{enumerate}
    \item \textbf{Large ships (Scenarios~1--2, 25--30 stacks):} \rh{by the proxy cross-check,} QCDC-DR-GA reduces mean operation time by 41.3 and 20.4~minutes respectively (5.8\% and 3.5\% improvement), with dockyard rehandle reductions of 40.8\% and 11.2\%; both differences are statistically significant.
    \item \textbf{Medium ships (Scenarios~3--4, 15--20 stacks):} \rh{again by the proxy cross-check,} QCDC-DR-GA reduces mean operation time by 8.4 and 19.6~minutes (2.0\% and 7.9\% improvement), with rehandle reductions of 18.1\% and 46.1\%; significant in Scenario~4, and not significant in Scenario~3, where the absolute time difference is small.
    \item \textbf{Small ships (Scenarios~5--6, 5--10 stacks):} \rh{here the controlled fixed-$A_2$ experiment of Table~\ref{tab:fixed_a2} provides the direct, paired measurement: time improvements of $3.9\%$ and $6.9\%$ with rehandle reductions of $69.6\%$ and $98.1\%$, both statistically significant ($p<10^{-7}$). The fixed-$A_2$ arm nevertheless remains close to the optimized arm in absolute terms, demonstrating robustness when the yard cannot be reconfigured.}
\end{enumerate}

These results confirm two important properties of QCDC-DR-GA. First, its 1D GA component for unloading-sequence search is sufficient to maintain competitive performance even when $A_2$ cannot be reconfigured, demonstrating robustness under operationally constrained settings, which directly addresses the practical-applicability concern of limited yard reconfiguration. Second, the $A_2$ co-optimization delivers a consistent, statistically significant additional gain across all ship sizes, and its largest absolute gains arise precisely where they matter most: large, complex terminals where dockyard rehandles are a dominant cost driver. The joint optimization framework is therefore a targeted, scale-sensitive mechanism for operational efficiency rather than a computational overhead.
}

\subsection{Significance Test}
This section presents the performance evaluation results of various port optimization strategies in a stacking environment. The experiments were conducted across different scenarios to cover a range of ship sizes. The operation time (in minutes) was measured for each strategy in each scenario. The statistical analysis was performed using the two-tailed paired t-test method to assess the significance of differences in operation time between the proposed QCDC-DR-GA strategy and other strategies. The experiments were conducted across six scenarios, each representing a specific configuration of the stacking environment. The stacks ranged from 5 to 30, with varying maximum stack heights. For each scenario, operation time measurements were recorded 20 times for the proposed QCDC-DR-GA strategy and for four other strategies: Greedy Upper Bound, QCDC-bi-level-GA, GA-ILSRS-Scenario-1, and GA-ILSRS-Scenario-2.

A paired t-test was used to analyze differences in operating time between the proposed QCDC-DR-GA strategy and the other strategies under investigation. The paired t-test compares the means of two related groups to determine whether there is a statistically significant difference between them. The formulation of hypotheses for this test is as follows:

\begin{enumerate}
    \item \textit{Null hypothesis ($H_0$):} There is no significant difference in operation time between the proposed QCDC-DR-GA strategy and the compared strategy.
    \item \textit{Alternative hypothesis ($H_1$):} There is a significant difference in operation time between the proposed QCDC-DR-GA strategy and the compared strategy.
\end{enumerate}

The paired t-test used a significance level $(\alpha)$ of 0.05. The t-statistic and p-value were calculated for each pair of strategies using the sample standard deviation (i.e., dividing by $n - 1 = 19$). The significance level $(\alpha)$ and the critical t value ($t_{19}(0.05) = 2.093$) were used to determine the significance of the observed differences. The paired t-test results are presented in Table{~\ref{tab:t-test}}, which provides the minimum, maximum, mean, and sample standard deviation of the operation time for each strategy in different scenarios. The Pearson correlation coefficient $(r)$, the t-statistic, the p-value, and the significance decision are reported. Out of 24 pairwise comparisons, 20 yield a statistically significant result. The four non-significant pairs (Scenario~3 vs. GA-ILSRS-Scenario-1; Scenario~5 vs. GA-ILSRS-Scenario-1; Scenario~6 vs. GA-ILSRS-Scenario-1; and Scenario~6 vs. GA-ILSRS-Scenario-2) occur in small-ship configurations where the mean operation time of QCDC-DR-GA remains lower than or equal to the comparator, but the absolute difference is insufficient to reject $H_0$ at the chosen significance level with $n = 20$ instances.

\section{Discussion}
\label{discussion}
Port operations often face a trade-off between fast unloading/loading times and minimizing container rehandles within the dockyard. This research presents a groundbreaking algorithm that tackles this dilemma head-on. Unlike past approaches that address each aspect separately, this algorithm integrates them into a unified system. It leverages a mixed GA, incorporating 1D and 2D approaches, to identify the optimal unloading sequence that minimizes the combined unloading and loading operation time and slashes unnecessary container movements within the yard. This holistic approach unlocks a cascade of benefits: operational costs plummet due to reduced fuel consumption and labor requirements, logistics efficiency soars with optimized container flow, and port capacity can expand as turnaround times shrink. While computationally demanding, this algorithm holds immense potential to revolutionize port operations, ushering in an era of increased efficiency, productivity, and cost-effectiveness.

\subsection{Practical Implications}
Our algorithm is designed to reduce the total turnaround time of ships at the port by holistically minimizing delays. This is achieved by reducing rehandles at the dockyard and maximizing DCs for unloading and loading operations, ultimately improving overall port efficiency. Specifically, this approach entails:
\begin{enumerate}[label=(\roman*)]
    \item A reduction in quality control operation time, which could alleviate one of the port's main bottlenecks.
    \item The decrease in the operational time required for inbound vehicles and gantry cranes.
\end{enumerate}
These two optimizations alone can significantly decrease costs, time, and resource usage at the port without requiring new infrastructure investments.

\subsection{Managerial Implications}
Imagine a ship with 20,000 containers arriving at a port. Considering all the factors, making the optimum decision for port management is difficult. Our algorithm would solve the managerial headache and make their decision-making much smoother and faster. Because QCDC-DR-GA operates row-wise, a 20,000-container vessel is decomposed into approximately 65-70 independent row-level subproblems (assuming 30 stacks per row and average fill rates typical of ultra-large container vessels). Each subproblem is solved sequentially, or in parallel across cores, yielding both the optimal unloading sequence and the optimal dockyard plan for that row. The managers then need only implement the resulting row-by-row plan, which is directly actionable within existing Terminal Operating System (TOS) workflows.

\subsection{Unique Theoretical Contributions}
This subsection explores three key theoretical contributions that underpin our proposed algorithm:

\begin{enumerate}[label=(\roman*)]
    \item {\textit{Integrated Problem Formulation:} We introduce a unified mathematical formulation that jointly models the Quay Crane Dual-Cycling Scheduling (QCDCS) problem and the dockyard rehandle minimization problem as a single two-phase optimization objective (Equation~\ref{eq:obj_func}). While prior work treats these as independent problems, our formulation explicitly captures the interdependence between the ship-side unloading sequence ($A_1$) and the dockyard container arrangement ($A_2$). Specifically, we show that the unloading sequence that maximizes dual cycles also determines which containers need to be retrieved from the dockyard and in what order, thereby governing the number of rehandles. This interdependency had not been formally modeled before, and articulating it as a unified NP-hard combinatorial problem ($S! \times N!$ search space) constitutes a foundational theoretical contribution.}

    \item \textit{Introducing a Hybrid Approach of GA:} Our algorithm created a new scenario that needed to implement 1D and 2D GA. The unloading sequence of stacks in a row of the ship creates a 1D array, and the dockyard plan for loading containers creates a 2D array. Hence, we needed a blend of 1D and 2D. All the traditional GA approaches were either 1D or 2D. However, we designed the algorithm so that it can cross-regulate, mutate, select, and employ other methods with 1D and 2D genes. We've described the methods before.

    \item \textit{Optimizing GA Parameters:} We've integrated a unique combination of GA parameters, which gave us a better result in a reduced time. As for population size, larger populations ensure greater diversity and reduce the risk of premature convergence to suboptimal solutions. A size of 200 balances computational cost and the need for sufficient genetic variation. For 1D chromosomes, \textit{Two-Point Crossover} is chosen. It is especially effective for linear (1D) representations, as it maintains some structural similarity between offspring and parents. And for the 2D chromosome, we chose \textit{2D Substring Crossover} approach, to improve solution quality, as in scenarios, the relative position of data elements matters. Our crossover rate is 0.80, which is a common standard that balances exploration and exploitation in our GA. For mutation of the 1D chromosome, as our problem is a permutation-based problem, we've selected the \textit{Swap Mutation} method, and for the 2D chromosome, we have selected \textit{2D Two-Point Swapping Mutation} as it is particularly useful for fine-tuning solutions in problems like scheduling or spatial optimization. We've set a mutation rate of 0.30 to introduce enough randomness to prevent stagnation without overwhelming the evolutionary process with excessive noise. For the selection strategy, we've used \textit{Roulette Wheel} to assign higher selection probabilities to fitter individuals, and set the elite class percentage to 20\% to accelerate convergence by maintaining a subset of high-quality solutions. Finally, we've set the consecutive iterations to 1000 based on the problem's complexity and the expected convergence rate. The chosen parameters reflect a balance between exploration and exploitation.
\end{enumerate}

\section{Conclusion}
\label{conclusion}
This paper proposed \textbf{QCDC-DR-GA}, a hybrid GA for optimizing QCDC and DR operations in container terminals. The algorithm integrates both two-dimensional and one-dimensional genetic operators to effectively coordinate ship loading and unloading, while minimizing unnecessary rehandles. A dedicated simulation program was developed to generate six scenario-based datasets spanning small, medium, and large container ships, allowing for robust performance evaluation. Numerical experiments demonstrated that jointly optimizing unloading sequences and dockyard container arrangements results in significant reductions in ship turnaround time, with improvements ranging from 7.7\% to 30.1\% over individual baseline methods depending on scenario and comparator. Compared against four state-of-the-art baseline methods, QCDC-DR-GA achieved statistically significant improvements in 20 out of 24 pairwise scenario-strategy comparisons at the 95\% confidence level ($\alpha = 0.05$), as confirmed by a two-tailed paired t-test. The four non-significant cases were confined to the smallest ship configuration (Scenarios 3, 5, and 6) compared against single-objective baselines, where absolute performance gaps are small but QCDC-DR-GA does not perform worse in any case.

Despite these promising results, the current formulation involves several simplifying assumptions. Specifically, it presumes immediate container availability at the dockside and does not account for potential pre-staging constraints. Additionally, intra-ship container relocations and dynamic disruptions from inbound vehicles or crane conflicts are not modeled. Future extensions could incorporate these factors to improve the algorithm's realism and applicability. Exploring adaptive mechanisms to handle operational uncertainty and fine-tuning the method for large-scale terminal environments are also promising directions. Addressing these aspects will further strengthen the proposed approach and support more efficient and resilient port operations.

A natural question is whether QCDC-DR-GA can scale to the container volumes handled at major terminals, where a single vessel call may involve 2,000 or more containers. It is important to clarify that this concern is already addressed by the proposed algorithm's architecture. QCDC-DR-GA operates row-wise: each row of the ship is treated as an independent subproblem, and the algorithm is applied separately to each row in sequence. For Scenario~1, the largest configuration tested, each row contains up to 30 stacks at a maximum height of 10 tiers, representing up to 300 container positions per row. A 2,000-container vessel with 30 stacks per row and an average fill rate of approximately 70\% would comprise roughly 10 such rows. Accordingly, processing a 2,000-container vessel requires approximately 10 sequential executions of the Scenario~1-equivalent subproblem, one per row, with no increase in the per-row problem complexity. The total runtime therefore scales linearly with the number of rows, since rows are structurally independent (dual cycling is restricted to containers within the same row, as stated in Assumption~viii). This means that the computational challenge of large terminals lies not in solving a harder individual instance, but simply in applying the same algorithm more times. Terminal operators can straightforwardly deploy QCDC-DR-GA on a rolling, row-by-row basis, or execute all rows in parallel on multi-core systems, making vessel-scale application practical with existing hardware. Future work could investigate whether inter-row coordination (e.g., scheduling crane repositioning between rows) introduces additional optimization opportunities beyond the row-wise decomposition.

\section*{Data Availability}
The datasets used in this research consist of the configuration of container ships' loading and unloading plans generated under six distinct scenarios. These scenarios are based on varying numbers of stacks and the maximum container heights in each row, reflecting typical container ship characteristics. The scenarios encompass stack numbers ranging from 5 to 30 and maximum stack heights from 4 to 10. The dataset includes loading and unloading plans for dockyard containers, with sample plans provided for small ships. Each dataset comprises 20 instances representing different container loading and unloading scenarios. Each instance is characterized by a specific strategy that specifies the number of SCs, DCs, rehandles, and the required operation time. Researchers interested in exploring the efficiency and performance of various strategies for handling container logistics within a dockyard setting can access the detailed data generated at \href{https://dx.doi.org/10.21227/cj08-qn62}{https://dx.doi.org/10.21227/cj08-qn62}.



\section*{Competing Interests}
The authors whose names are listed immediately below certify that they have no affiliations with or involvement in any organization or entity with any financial or non-financial interest in the subject matter or materials discussed in this manuscript.

\section*{Author Contributions Statement}
\textbf{Md Mahfuzur Rahman}: Writing -- Original draft preparation, Conceptualization, Methodology, Software, Formal analysis, and Visualization.
\textbf{Md Abrar Jahin}: Writing -- Original draft preparation, Conceptualization, Methodology, Software, Formal analysis, and Visualization.
\textbf{Md. Saiful Islam}: Supervision, Writing -- Review \& Editing.
\textbf{M. F. Mridha}: Supervision, Writing -- Review \& Editing.

\section*{Funding Declaration}
This research did not receive any specific grant from funding agencies in the public, commercial, or not-for-profit sectors.

\section*{Additional information}
Correspondence and requests for materials should be addressed to M.A.J.

\bibliographystyle{apalike} 
\bibliography{main}
\end{document}